%% file: main.tex
\documentclass[sigplan,screen]{acmart}

\usepackage[]{hyperref}
    
\AtBeginDocument{%
  }

\usepackage{enumitem}
\usepackage{etoolbox,xspace}
\usepackage{algorithm}
\usepackage[noend]{algorithmic}
\usepackage{amsmath}
\usepackage{caption}
\usepackage{subcaption}
\usepackage{multirow}
\usepackage{array}
\usepackage{booktabs}
\usepackage{balance}

\newcommand{\sys}{\texttt{llm.npu}\xspace}

\newcommand{\revision}[1]{#1}

\copyrightyear{2025}
\acmYear{2025}
\setcopyright{acmlicensed}
\acmConference[ASPLOS '25] {Proceedings of the 30th ACM International Conference on Architectural Support for Programming Languages and Operating Systems, Volume 1}{March 30--April 3, 2025}{Rotterdam, Netherlands.}
\acmBooktitle{Proceedings of the 30th ACM International Conference on Architectural Support for Programming Languages and Operating Systems, Volume 1 (ASPLOS '25), March 30--April 3, 2025, Rotterdam, Netherlands}
\acmISBN{979-8-4007-0698-1/25/03}
\acmDOI{10.1145/3669940.3707239}

\begin{document}

\title{Fast On-device LLM Inference with NPUs}

\input{sec-author}

    

\begin{abstract}
On-device inference for Large Language Models (LLMs), driven by increasing privacy concerns and advancements of mobile-sized models, has gained significant interest.
However, even mobile-sized LLMs (e.g., Gemma-2B) encounter unacceptably high inference latency, often bottlenecked by the prefill stage in tasks like screen UI understanding.

We present \sys, the first LLM inference system utilizing on-device Neural Processing Unit (NPU) offloading to reduce prefill latency.
\sys enhances NPU offloading efficiency by re-constructing the prompt and model in three levels:
(1) At \textit{prompt level}, it divides variable-length prompts into multiple fixed-sized chunks while maintaining data dependencies;
(2) At \textit{tensor level}, it identifies and extracts significant outliers to run on the CPU/GPU in parallel with minimal overhead;
(3) At \textit{block level}, it schedules Transformer blocks in an out-of-order manner to the CPU/GPU and NPU based on their hardware affinity and sensitivity to accuracy.
Compared to competitive baselines, \sys achieves 22.4$\times$ faster prefill speed and 30.7$\times$ energy savings on average, and up to 32.8$\times$ speedup in an end-to-end real-world application.
For the first time, \sys achieves more than 1,000 tokens/sec prefilling for a billion-sized model.
\end{abstract}



\maketitle




\input{sec-intro}
\input{sec-background}

\input{sec-design}
\input{sec-eval}

\input{sec-discuss}

\input{sec-related}

\input{sec-conclusion}

\input{sec-ack}

\input{sec-ae}

\bibliographystyle{plain}
\balance
\bibliography{acmart}


\end{document}

%% file: sec-author.tex
\author[Daliang Xu]{Daliang Xu}
\affiliation {
	\institution{Key Lab of HCST (PKU), MOE; SCS, Peking University, China
}
	\country{}
}
\email{xudaliang@pku.edu.cn}

\author[Hao Zhang]{Hao Zhang}
\affiliation {
	\institution{Beijing University of Posts and Telecommunications, China
}
	\country{}
}
\email{hao\_zhang@bupt.edu.cn}

\author[Liming Yang]{Liming Yang}
\affiliation {
	\institution{Key Lab of HCST (PKU), MOE; SCS, Peking University, China
}
	\country{}
}
\email{yangliming2018@pku.edu.cn}

\author[Ruiqi Liu]{Ruiqi Liu}
\affiliation {
	\institution{Key Lab of HCST (PKU), MOE; SCS, Peking University, China
}
	\country{}
}
\email{rich@stu.pku.edu.cn}

\author[Gang Huang]{Gang Huang}
\affiliation {
	\institution{Key Lab of HCST (PKU), MOE; SCS, Peking University, China
}   \institution{National Key Laboratory of Data Space Technology and System, China
}
	\country{}
}
\email{hg@pku.edu.cn}

\author[Mengwei Xu]{Mengwei Xu$^*$}
\affiliation {
	\institution{Beijing University of Posts and Telecommunications, China
}
	\country{}
}
\email{mwx@bupt.edu.cn}

\author[Xuanzhe Liu]{Xuanzhe Liu$^*$}
\affiliation {
	\institution{Key Lab of HCST (PKU), MOE; SCS, Peking University, China
}
	\country{}
}
\email{liuxuanzhe@pku.edu.cn}
\thanks{$^*$Corresponding authors.}

\renewcommand{\shortauthors}{Daliang Xu et al.}

%% file: sec-intro.tex
\section{Introduction}






With rising privacy concerns~\cite{gdpr}, there is growing interest in running Large Language Models (LLMs) locally on mobile devices, known as \textit{on-device LLMs}~\cite{yuan2024mobile,yin2024elms,yin2024llm}, e.g, Apple Intelligence~\cite{apple-intelligence} and Android AI Core~\cite{AI-Core}.
Meanwhile, the advancement of mobile-sized language models (1B–10B parameters), such as Qwen2-1.5B and Phi3-3.7B, has demonstrated their ability to perform comparably to significantly larger models like GPT-3, despite having a reduced parameter count~\cite{abdin2024phi,qwen2,MMLU-leader-board,lu2024small}. This progress makes the deployment of on-device language models feasible.
Without giving away private data, on-device LLM inference catalyzes novel mobile applications~\cite{xu2019first}, such as UI task automation~\cite{li2024personal} (e.g., translating users' language commands into UI operations such as ``forward the unread emails to Alice'')
and automated message reply~\cite{email}.

However, the high inference latency remains as a significant obstacle to practical on-device LLMs.
To accomplish a UI task, LLM needs to ingest the screen view hierarchy (typically 600-800 tokens~\cite{wen2024autodroid,LlamaTouch}) to generate corresponding UI operations step by step~\cite{wen2023empowering}.
As will be shown in $\S$\ref{sec:bg-llm-analysis}, each such step takes 8.1 seconds for Qwen1.5-1.8B~\cite{qwen2}, and thereby more than 40 seconds to finish a 5-step UI task.
Similarly, the Gemma-2B model requires 26.7 seconds to automatically reply to an email by mimicking the user's tone based on historical email data (with 1500 tokens).
Diving into those tasks, we find out that the prompt processing (prefill stage) often dominates the end-to-end inference latency, e.g., 94.4\%--98.8\% for UI automation tasks.
This is because on-device LLM tasks often involve long-context understanding for handle personalized tasks.
Unfortunately, existing research efforts primarily focus on accelerating the text generation speed (decoding stage), such as activation sparsity~\cite{xue2024powerinfer,song2023powerinfer}
and speculative decoding~\cite{kim2023big, miao2023specinfer, yang2023predictive}.
Therefore, this work mainly targets improving the prefill speed of on-device LLMs.

LLM prefilling is compute-bounded~\cite{zhong2024distserve,patel2023splitwise,wu2024loongserve};
yet, mobile CPU and GPU have limited parallel computing capacity~\cite{yi2020heimdall,han2024pantheon}.
Instead, we are motivated by a key opportunity that Neural Processing Units (NPUs) are ubiquitously available in modern mobile devices, e.g., Qualcomm Hexagon NPU and Google Edge TPU.
These mobile NPUs are efficient at integer vector operations, delivering computing capability up to 73 TOPS~\cite{8gen3}.
On CNNs, their improvements over mobile CPU/GPUs are demonstrated to be up to 18$\times$/4$\times$, respectively~\cite{xu2022mandheling}. 
Mobile NPUs are also more energy-efficient, and have less workloads contention as compared to CPU/GPU.

Surprisingly, with such promised advantages, there exists no systems supporting LLM inference on COTS mobile NPUs.
Indeed, our preliminary efforts show that directly employing mobile NPUs for LLM inference does not offer performance benefits due to the following challenges.

$\bullet$ \textit{Costly preparation for variable-length prompts.}
Mobile NPUs typically support only inference on static shapes, while LLM prompt length is dynamic (with a max context length).
Re-preparing and optimizing the LLM execution graph on NPUs for each different-sized prompt is costly on mobile devices (e.g., 11 seconds for the Gemma-2B model).
On the other hand, simply padding the LLM request to the same max context length wastes precious computing resources.

$\bullet$ \textit{Mismatch between LLM quantization algorithms and mobile NPU design.}
Due to the presence of outlier activations~\cite{xiao2023smoothquant,dettmers2022gpt3}, state-of-the-art LLM quantization methods often use per-group quantization to maintain high accuracy.
It partitions original activations and weights tensors into multiple groups with independent quantization scales to avoid the impacts of outliers on others.
However, our investigation shows that mobile NPUs cannot perform per-group MatMul directly (Table~\ref{tab:sec-bg-npus}). Instead, they must split the MatMul into multiple group-sized sub-tensor MatMuls and then reduce the sub-tensor intermediate results using a float sum operation. This process hampers NPU efficiency and incurs up to 10.7$\times$ performance overhead.

$\bullet$ \textit{Floating point (FP) operations cannot be eliminated.}
Mobile NPUs generally provide significant integer-based MatMul acceleration but are weak at FP operations.
However, LLMs can be hardly quantized into integer-only execution with minimal accuracy loss.
Existing quantized LLMs still rely on float operators like \texttt{LayerNorm} and \texttt{Attention}.
Scheduling those FP operators out of NPU easily increases the inference critical path.

This work presents \sys, the first LLM inference system with efficient on-device NPU offloading. 
The primary design goal of \sys is to reduce the prefill latency and energy consumption.
It targets the mainstream decoder-only transformer architecture of LLMs (e.g., LlaMA, GPT, etc). 
The key idea is to maximize prefill execution on mobile NPUs to accelerate integer computation while keeping essential float operations on the CPU/GPU to maintain accuracy.
To overcome the aforementioned challenges and enhance NPU offloading efficiency, \sys re-constructs the prompt and model at three levels:
(1) At \textit{prompt level:} \sys divides variable-length prompts into multiple fixed-sized chunks while maintaining data dependencies;
(2) At \textit{tensor level:} \sys identifies and extracts significant outliers to run on the CPU/GPU;
(3) At \textit{block level:} \sys schedules Transformer blocks to the CPU/GPU and NPU based on their hardware affinity and sensitivity to accuracy.
The corresponding novel techniques are detailed as follows:

$\bullet$ \textit{Chunk-sharing graphs}
($\S$\ref{sec:design-chunk})
\sys splits variable-length prompts into fixed-sized ``chunks'' with pre-built subgraphs, reducing graph preparation time.
Each variable-length prompt is then executed on those graphs with chunk-level causal dependency.
This approach leverages the insight that the token generation depends only on preceding tokens in the decoder-only LLMs.
However, loading multiple pre-built chunk graphs simultaneously incurs significant memory overhead, e.g., 2--4$\times$ more than the LLM weights.
To address this, \sys further identifies operators that are irrelevant to prompt size (e.g., FFN) and shares them across each chunk execution, reducing the memory overhead by up to 4$\times$.

$\bullet$ \textit{Shadow outlier execution}
($\S$\ref{sec:design-outlier})
addresses the activation outlier problem without compromising NPU efficiency.
Without changing per-tensor MatMul on NPU, \sys extracts the activation outlier channels into a compact tensor and executes it on the CPU/GPU in parallel,  since outliers are very sparse  (0.1\%--0.3\% of total channels).
This design further raises issues of increased memory footprint due to duplicating MatMul weights in CPU memory and synchronization overhead between CPU/GPU and NPU.
Therefore, based on the observation that outliers are more likely to appear at a small set of channel positions, \sys optimizes memory usage by only keeping those ``hot channels'' weights in memory, and retrieves others from disk on demand.
\sys also prunes unimportant outliers at layer-level by measuring outliers' importance to reduce the synchronization overhead.


$\bullet$ \textit{Out-of-order subgraph execution}
($\S$\ref{sec:design-balance-execution})
\sys is guided by a key insight that multiple subgraphs can be scheduled in an out-of-order manner, without strictly following the chunk sequence in the original prompt.
This significantly enlarges the scheduling space of \sys to minimize the execution bubbles brought by CPU/GPU float operation.
Given that finding the optimal out-of-order execution order is an NP-hard problem, \sys employs a microsecond-level online scheduling algorithm. This algorithm is based on the observation that the workload of the NPU is heavier and constitutes the critical path. Therefore, when selecting which subgraph to execute, \sys prioritizes those subgraphs having a more significant impact on reducing NPU stalls, rather than focusing solely on the execution latency of the subgraph.
Notably, \sys's scheduling algorithm does not maximize parallel processing capability. Instead, \sys aims to maximize the utilization of NPUs while minimizing the impact of CPU/GPU workloads.

\textbf{Implementation and evaluations.}
We have implemented \sys on top of MLLM~\cite{mllm} and QNN~\cite{QNN},
with 10K lines of C/C++ and assembly code.
We evaluated \sys with five mobile-sized LLMs (Qwen1.5-1.8B~\cite{qwen2}, Gemma-2B~\cite{gemma2b}, phi2-2.7B~\cite{phi2-2.7}, LlaMA-2-7B~\cite{Llama-2-7b}, and Mistrial-7B~\cite{Mistral-7B}), four LLM benchmarks,
and two mobile devices (Xiaomi 14 and Redmi K60 Pro).
We compared \sys with five competitive baselines, including three industrial open-source engines (llama.cpp~\cite{llama-cpp}, TFLite~\cite{tflite}, MNN~\cite{mnn}) and two state-of-the-art research prototypes (MLC-LLM~\cite{mlc-llm} and PowerInfer-v2~\cite{xue2024powerinfer}).
The experiments show that \sys significantly and consistently outperforms all baselines in terms of prefill latency and energy consumption while preserving inference accuracy (<1\% loss compared to FP16).
It is 7.3$\times$--18.4$\times$ faster than baselines on CPU, and 1.3$\times$--43.6$\times$ on GPU with a prompt length of 1024.
It also achieves a 1.9$\times$--59.5$\times$ energy reduction.
To our best knowledge, \sys is the first system that achieves $>$1000 tokens/sec of prefill speed on COTS mobile devices for billion-sized LLMs. 
In end-to-end real-world applications, \sys reduces the inference latency (prefill+decode) by 1.4$\times$--32.8$\times$ compared to the baselines.

\textbf{Contributions} are summarized as follows:
\begin{itemize}[leftmargin=10pt]
    \item We thoroughly investigate the challenges and opportunities of accelerating LLM prefilling with mobile NPUs.
    \item We present the first LLM inference engine with efficient mobile NPU offloading, featuring three novel techniques: chunk-sharing graph, shadow outlier execution, and out-of-order subgraph execution.
    \item We perform comprehensive experiments on \sys that demonstrate its superior performance over competitive baselines. \revision{The code of \sys is available at \url{https://github.com/UbiquitousLearning/mllm}.}
\end{itemize}

%% file: sec-background.tex
\section{Background}

\subsection{On-device LLM Inference Analysis} \label{sec:bg-llm-analysis}




On-device LLMs are increasingly used in cutting-edge scenarios such as Apple Intelligence~\cite{apple-intelligence}, UI automation~\cite{wen2023empowering}, and automated email reply~\cite{email},  due to the enhanced privacy protections. To empower these applications, many lightweight LLMs have been developed, as summarized in Table~\ref{tab:sec-bg-mllm-max-tokens}.
For instance, the Qwen1.5-1.8B model on the llama.cpp exhibits delays of 8.1 seconds for one-step UI automation, 21.7 seconds for automated email replies, and 8.4 seconds for chat summary on average using mobile CPUs, as evaluated on the DroidTask~\cite{wen2024autodroid,LlamaTouch}, LongBench datasets~\cite{bai2023longbench} and Persona-Chat~\cite{Persona-Chat}, which is impractical for real-world deployment.
However, their inference latency remains a significant challenge.

\revision{To explore this issue in detail, we break down the inference procedure discussed above. 
The results are summarized in Figure~\ref{fig:sec-bg-inference-prefill-decoding-compare}.
Surprisingly, we find that compared to the decoding phase that has been extensively studied, the prefill stage is often the bottleneck in typical mobile applications. For example, the prefill stage constitutes 88.3\%--98.8\% of the total latency on a mobile CPU for UI task automation, chat summaries, and context-aware generation tasks. Even when using mobile GPUs, the prefill phase remains predominant, accounting for 54.2\% to 91.7\% of the total latency.}
As the prompt length increases, the prefill stage's proportion of the total inference time also rises.

Several factors contribute to this situation:
(1) Mobile CPUs/GPUs lack the parallelism capabilities of cloud GPUs~\cite{yi2020heimdall,han2024pantheon}, being primarily designed for handling application logic or rendering tasks.
(2) Mobile LLM tasks often require long prompts for personalized, context-aware generation. \revision{For instance, automated email reply often requires extensive user data in prompt, such as historical emails, schedules, and location information, amounting to 1,168--1,835 tokens in the LongBench dataset~\cite{bai2023longbench}. Similarly, on-device LLMs engaged in UI automation must process a substantial volume of UI annotation tokens (XML or HTML) and user commands, ranging from 505--827 tokens in the DroidTask dataset~\cite{wen2024autodroid,LlamaTouch}. Yet, the output lengths for these applications are typically short, averaging 7.9 tokens for the LongBench dataset and 3.5 tokens for the DroidTask dataset. In the context of chat summaries, such as the Persona-Chat dataset~\cite{Persona-Chat}, the prompt and output lengths are relatively balanced, with input sizes between 488--584 tokens and an average output of 44 tokens.
}
(3) Mobile LLMs now support long context windows.
For instance, recent models like Qwen2-1.5B can accommodate context windows of up to 32K tokens, as illustrated in Table~\ref{tab:sec-bg-mllm-max-tokens}.

\input{table-bg-mllm-max-tokens}


\input{figure-inference-prefill-decoding-compare}

\subsection{Opportunity: Mobile NPUs} \label{sec-bg-npu}






To optimize prefill latency, \sys leverages a key opportunity: modern mobile SoCs ubiquitously include mobile neural processing units (NPUs) that are well-suited for integer operations, such as INT8-based matrix multiplication. Table~\ref{tab:sec-bg-npus} summarizes the specifications of well-known mobile NPUs provided by mainstream vendors. For example, Qualcomm's mobile SoCs feature Hexagon NPUs, achieving up to 73 trillion INT8 operations per second. According to AI-Benchmark~\cite{ai-benchmark}, the Hexagon NPU in the Xiaomi 14 can infer the MobileNet-V2 model in just 0.6 ms, 23$\times$ faster than a mobile CPU and 3.2$\times$ faster than a mobile GPU.

\input{table-NPUs}

\textbf{Mobile NPU architecture.} Mobile NPUs deliver significant performance benefits by single instruction multiple data (SIMD) architecture.
For instance, Hexagon NPUs support 1024-bit INT8 vector arithmetic, allowing multiple SIMD instructions to execute in parallel. However, their floating-point computation capabilities are relatively weak compared to mobile GPUs. 
With clock frequencies between 500 and 750 MHz, mobile NPUs are more energy-efficient than mobile CPUs and GPUs. Additionally, unlike cloud GPUs that have separate physical memory, mobile NPUs are integrated within mobile SoCs, sharing the same physical memory with mobile CPUs, eliminating the need for memory copying during NPU execution.

\input{table-bg-matmul-npu}
\textbf{Micro benchmark comparison.}
\revision{To evaluate the performance of INT8 MatMul on mobile NPUs, we conducted preliminary experiments on the Redmi K70 Pro using MatMul sizes commonly used in mobile LLMs, as summarized in Table~\ref{tab:sec-bg-matmul-npu}. INT8 MatMul on mobile NPUs achieved a 4.5--5.8$\times$ speedup compared to CPU INT8 and a 1.8--3.5$\times$ improvement over GPU FP16. The performance gains increase with larger computational workloads. However, performing FP16 MatMul on the mobile NPU resulted in performance up to 159$\times$ slower than CPU INT8. These results align with the INT8 SIMD architecture of mobile NPUs, confirming that mobile NPUs are best suited for accelerating INT8 matrix multiplication.}

\input{figure-bg-NPU-programming-workflow}

\textbf{DNN execution workflow on mobile NPUs.}
Executing DNNs on mobile NPUs involves configuring the NPU environment, creating the compute graph, optimizing the graph, executing the graph, and freeing the graph, as shown in Figure~\ref{fig:sec-bg-npu-programming}.
Typically,  creating and optimizing the compute graph are most time-consuming, where the former includes translating models into the NPU-required intermediate representation, and memory allocation, that takes 300--500ms, while the latter includes optimization for adjusting memory layout, execution order, and operator fusion, taking many seconds.
In addition, the closed-source nature of NPU SDKs limits further adaptation for LLMs.

\subsection{Gaps between LLMs and Mobile NPUs} \label{sec-bg-observations}
Given its inherent advantages, we are surprised to find that none of existing DNN engines support LLM acceleration on mobile NPUs.
We then dig into the underlying reasons and find a huge gap between existing mobile NPUs design and LLM inference pipeline.
\noindent $\bullet$ \textbf{LLM prefill phase relies on variable-length prompts, leading to excessive time spent on building and compiling the NPU graph.}
As illustrated in Figure~\ref{fig:sec-bg-npu-programming}, before the compute graph can be executed on the mobile NPU, it must be built and optimized, a process taking tens of seconds. For instance, building the graph for the Gemma 2B model using QNN framework takes 360 ms, and graph optimization requires 11.54 seconds. Unlike CNN models, which are built and optimized once and can be executed multiple times with the same input shape, the LLM prefill phase must handle variable-length prompts, necessitating rebuilding and re-optimization of compute graphs for each inference. Consequently, using mobile NPUs in this scenario offers no performance benefit and is often slower than using a CPU.



\input{figure-bg-quantization-approach}

\input{figure-bg-quantization-accuracy-speed}

\noindent $\bullet$ \textbf{The existence of activation outliers makes LLM difficult to quantize at whole-tensor level, yet a more fine-grained group-level quantization hampers NPU efficiency.}
Our preliminary experiments, shown in Figure~\ref{fig:sec-bg-quantization-inference-accuracy-speed}, indicate that two popular quantization algorithms (K-Quant~\cite{llama-cpp} and AWQ~\cite{lin2023awq}) incur significant inference overhead by 8.1$\times$--10.7$\times$, as compared to per-tensor quantization (Figure~\ref{fig:sec-bg-quantization-approach}(a)).
This is because algorithms like K-Quant and AWQ use fine-grained \textit{per-group quantization} (Figure~\ref{fig:sec-bg-quantization-approach}(b)) to maintain high accuracy. These algorithms divide activations and weights into multiple groups, each with an independent quantization scale. On NPUs, this approach requires dividing the MatMul operation into several sub-tensor MatMuls, which fails to fully leverage the capabilities of mobile NPUs. Additionally, it necessitates aggregating intermediate results with floating-point additions, resulting in extra floating-point computations. The exception is SmoothQuant~\cite{xiao2023smoothquant}, which uses per-tensor quantization but suffers from significant accuracy loss, such as a 3.9\% and 8.4\% drop on the HelloSwag dataset for the LlaMA-2-7B and Qwen1.5-1.8B model, respectively.


\input{figure-bg-llm-inference}
\input{table-bg-module-format}

\noindent $\bullet$ \textbf{On-device LLM inference relies on floating-point operations, conflicting with the NPU's design for INT8 acceleration.} 
Figure~\ref{fig:sec-bg-llm-inference} illustrates a typical workflow for quantized inference.
To ensure inference accuracy, only linear layers (highlighted in blue) perform matrix multiplication in the INT8/INT4 data format. For these layers, the activation $x$ is quantized to INT8/INT4 before performing the dot product with weights. Other operations, such as Attention and LayerNorm (highlighted in orange), are computed in floating-point format. 
Table~\ref{tab:sec-bg-module-format} further summarize the operator data formats in state-of-the-art quantization inference algorithms. All of them depend on float Attention and normalization operations. Given that float operations dramatically degrade performance on mobile NPUs, these operations cannot be efficiently executed on the NPU with significant overhead.


%% file: table-bg-mllm-max-tokens.tex
\begin{table}[t]
\centering
\caption{The max context length for mobile-sized LLMs.
}
\resizebox{0.46\textwidth}{!}{%
\begin{tabular}{lrrlrr}
\hline
\textbf{Model} & \textbf{Max Context} & \textbf{Year} & \textbf{Model} & \textbf{Max Context} & \textbf{Year} \\ \hline
Opt-1.3B       & 2K                   & 2022.5        & TinyLLaMA-1.1B & 2K                   & 2023.9        \\ \hline
StableLLM-3B     & 4K                   & 2023.10       & phi-2-2.7B     & 2K                   & 2023.12       \\ \hline
Gemma-2B       & 8K                   & 2024.2        & Qwen1.5-1.8B   & 32K                  & 2024.2        \\ \hline
Phi3-mini-3.8B & 128K                 & 2024.5        & Qwen2-1.5B     & 32K                 & 2024.6        \\ \hline
\end{tabular}%
}
\label{tab:sec-bg-mllm-max-tokens}
\end{table}

%% file: figure-inference-prefill-decoding-compare.tex
\begin{figure}[t]
    \centering
    \begin{subfigure}[t]{0.23\textwidth}
        \includegraphics[width=\textwidth]{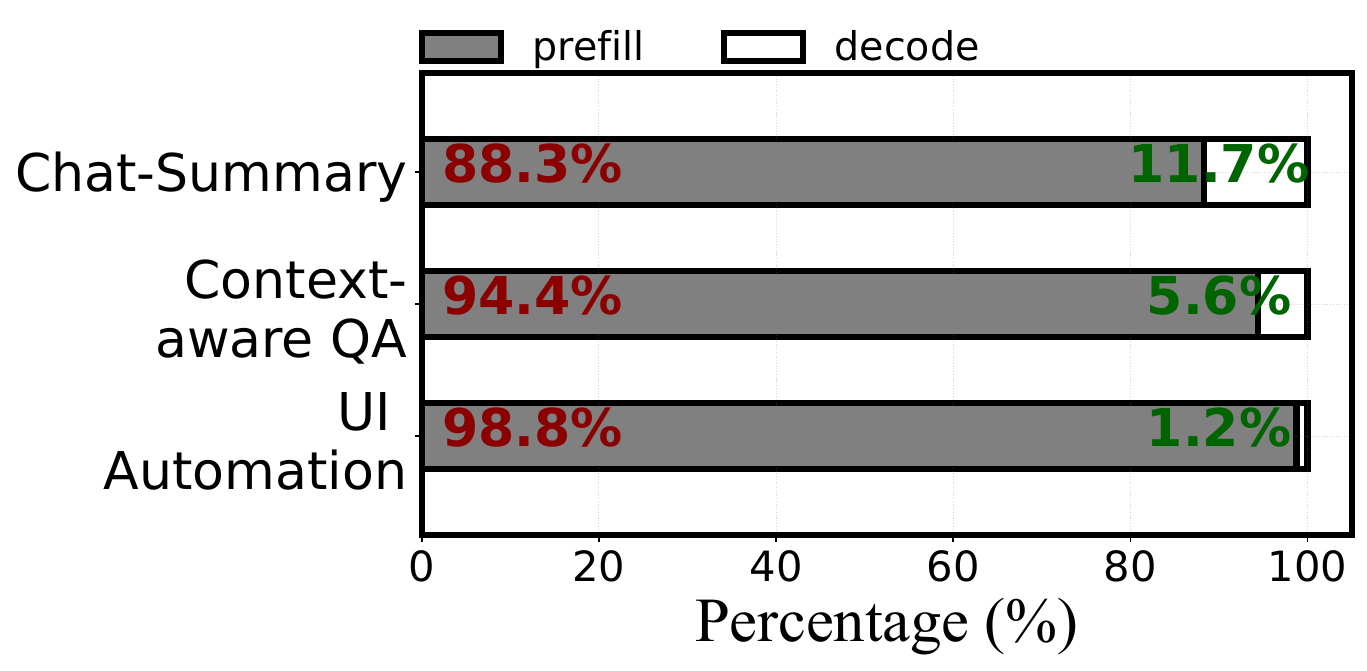}
        \caption{CPU results}
    \end{subfigure}
    \begin{subfigure}[t]{0.23\textwidth}
        \includegraphics[width=\textwidth]{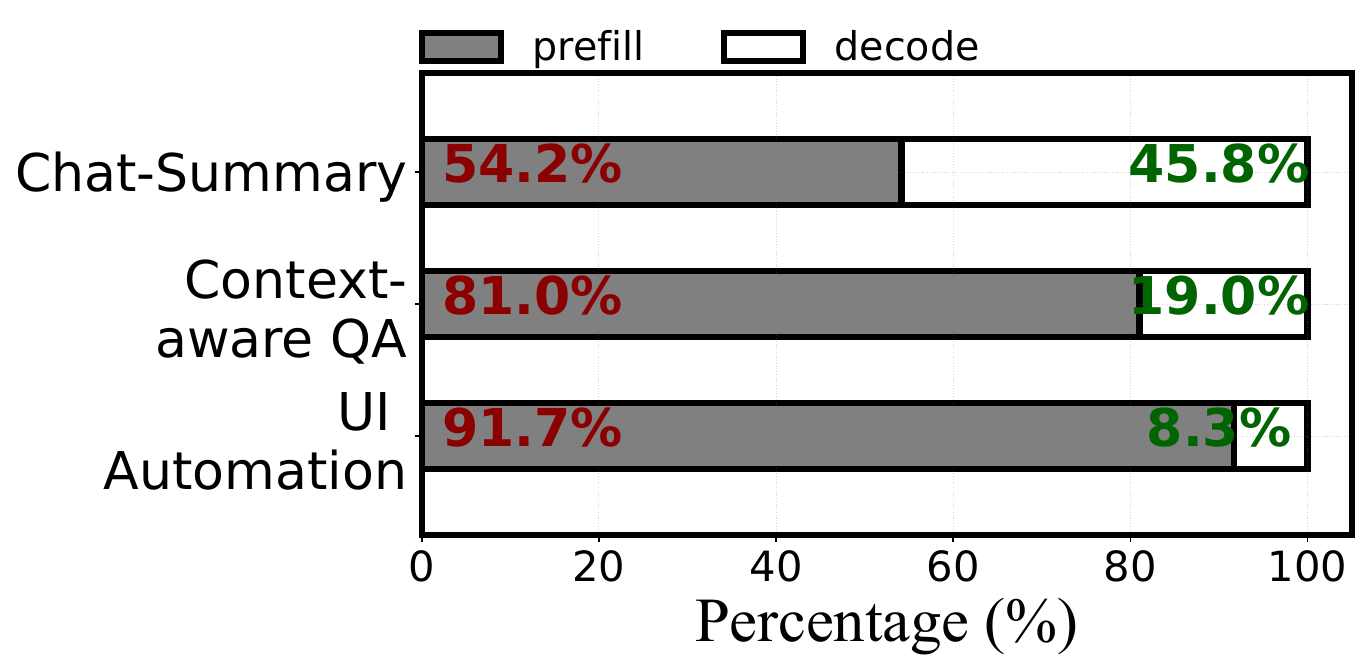}
        \caption{GPU results}
    \end{subfigure}
    \caption{\revision{Breakdown of end-to-end inference latency for UI automation, context-aware QA, and chat summaries. CPU evaluation uses llama.cpp as the on-device inference engine, while GPU evaluation uses TFLite as the simulation engine.}}
    \label{fig:sec-bg-inference-prefill-decoding-compare}
\end{figure}

%% file: table-NPUs.tex
\begin{table}[t]
\centering
\caption{Specifications of well-known mobile NPUs provided by mainstream vendors.}
\small
\resizebox{0.48\textwidth}{!}{%
\begin{tabular}{lllrrr}
\hline
\textbf{Vendor} & \textbf{Latest NPU} & \textbf{SDK} & \textbf{Open} & \textbf{Group} & \textbf{INT8 Perf.} \\ \hline
Qualcomm        &     Hexagon NPU~\cite{8gen3}          & QNN~\cite{QNN}  &  $\times$ &   $\times$    &       73 TOPS        \\ \hline
Google          &     Edge TPU~\cite{edgetpu}          &       Edge TPU API~\cite{Edge-TPU-API} &  $\times$    &  $\times$     &   4 TOPS            \\ \hline
MediaTek        &     MediaTek APU 790~\cite{APU-790}  &    NeuroPilot~\cite{neuropilot} &  $\times$    &  N/A   &    60 TOPS           \\ \hline
Huawei          &      Ascend NPU~\cite{Ascend-NPU}        &   HiAI~\cite{hiai} &  $\times$  & $\times$    &  16 TOPS                   \\ \hline
\multicolumn{6}{l}{\begin{tabular}[l]{@{}l@{}}"Open": Open-source?; "Group": Support per-group quantization MatMul? "N/A": No available \\ documents for public; "INT8 Perf.": Int8 performance.\end{tabular}}
\end{tabular}%
}
\label{tab:sec-bg-npus}
\end{table}

%% file: table-bg-matmul-npu.tex
\begin{table}[t]
\centering
\caption{\revision{Execution latency (ms) for various MatMul sizes on Redmi K70 Pro.}}
\resizebox{0.48\textwidth}{!}{%
\begin{tabular}{rrrrrr}
\hline
\textbf{Matrix A} & \textbf{Matrix B} & \textbf{NPU INT8} & \textbf{CPU INT8} & \textbf{GPU FP16} & \textbf{NPU FP16} \\ \hline
64*2048  & 2048*2048  &     0.9        &      4.2 (4.6$\times$)             &    1.7 (1.9$\times$)     &  252 (193$\times$)  \\ \hline
64*2048  & 2048*8192  &        1.5           &   6.8 (4.5$\times$)             &       4.8 (3.2$\times$)   &    986 (657$\times$)      \\ \hline
64*2048 & 2048*11008 &        2.0         &    11.6 (5.8$\times$)             &     6.9 (3.5$\times$)      &   1207 (603$\times$)       \\ \hline
32*4096  & 4096*4096  &     1.7       &    7.5 (4.4$\times$)   &       3.1 (1.8$\times$)   &    1054 (620$\times$)   \\ \hline
32*4096  & 4096*8192  &      2.9        &   13.1 (4.5$\times$)               &     7.7 (2.6$\times$)      &     2009 (692$\times$)     \\ \hline
32*4096 & 4096*11008 &       4.1         &   19.6 (4.8$\times$)        &    10.4 (2.5$\times$)     &    3112 (759$\times$)      \\ \hline
\end{tabular}%
}
\label{tab:sec-bg-matmul-npu}
\end{table}

%% file: figure-bg-NPU-programming-workflow.tex
\begin{figure}[t]
    \centering
    \includegraphics[width=0.45\textwidth]{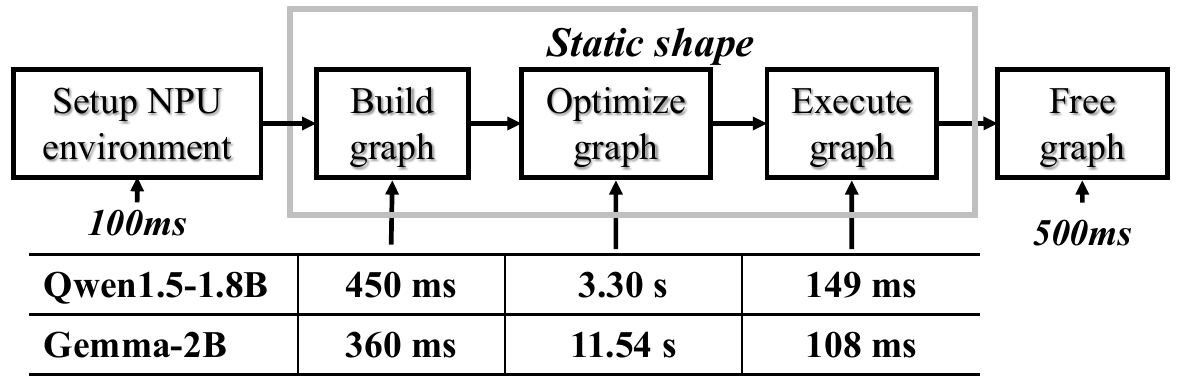}
    \caption{The workflow of executing DNNs on mobile NPUs, with latencies for each procedure on QNN~\cite{QNN}.}
    \label{fig:sec-bg-npu-programming}
\end{figure}

%% file: figure-bg-quantization-approach.tex
\begin{figure}[t]
    \centering
    \includegraphics[width=0.45\textwidth]{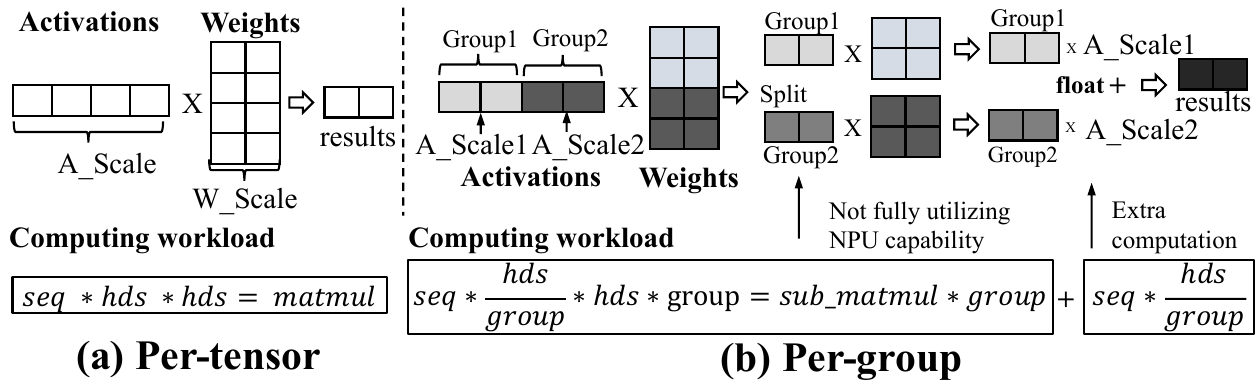}
    \caption{Per-tensor quantization MatMul and per-group quantization MatMul. $seq$, $hds$, $group$ represent sequence length, hidden size, and group number, respectively.}
    \label{fig:sec-bg-quantization-approach}
\end{figure}

%% file: figure-bg-quantization-accuracy-speed.tex
\begin{figure}[t]
    \centering
    \includegraphics[width=0.21\textwidth]{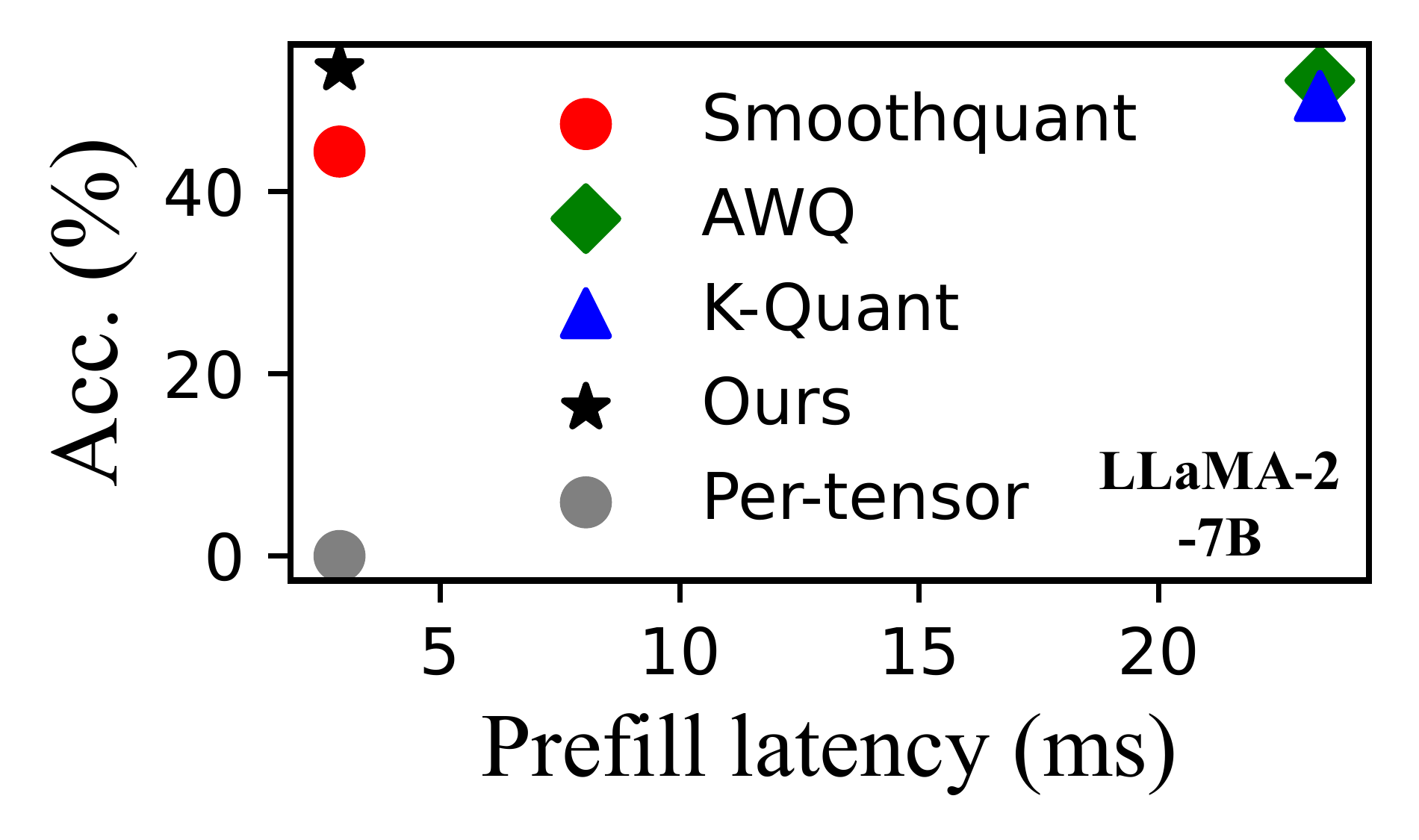}
    \includegraphics[width=0.21\textwidth]{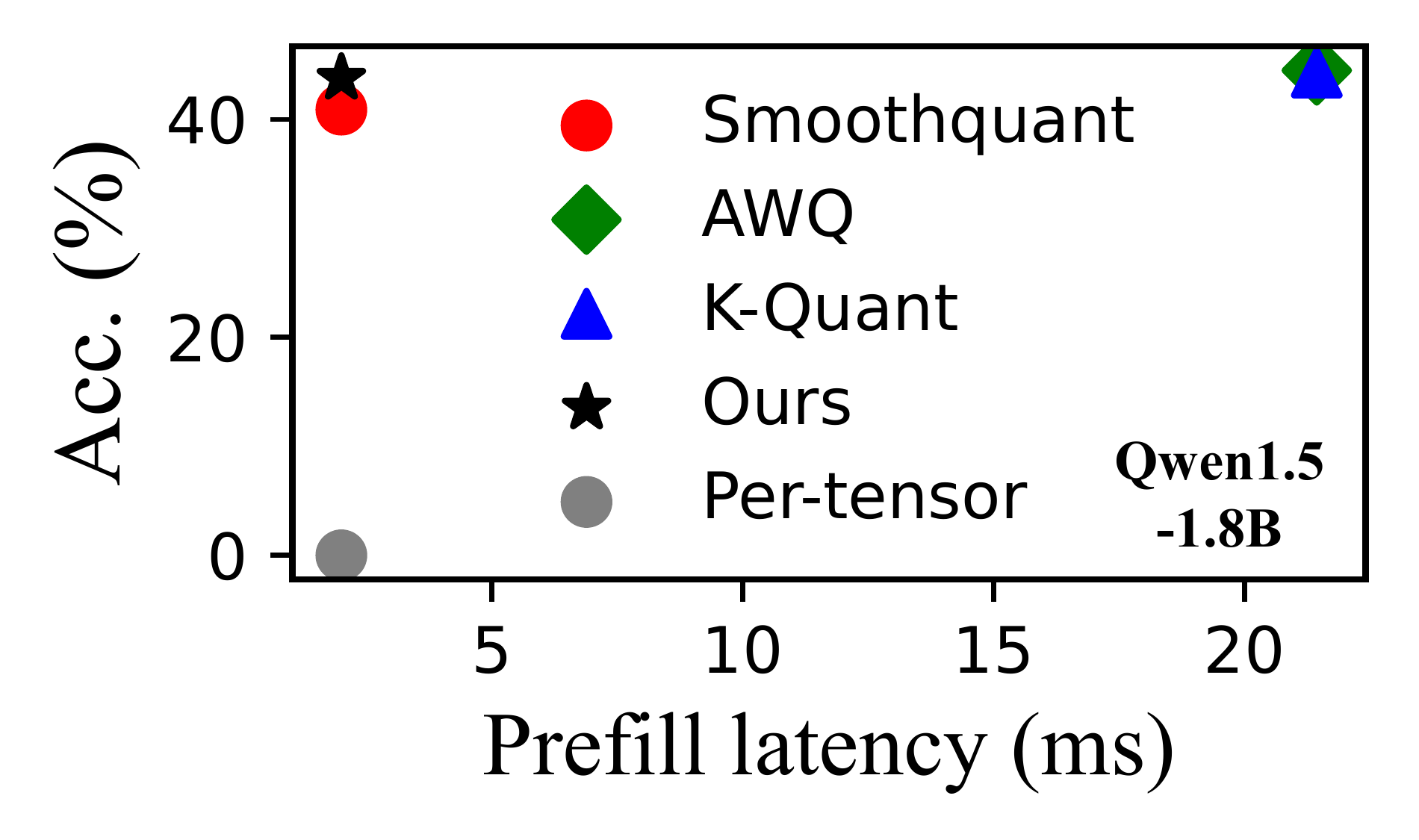}
    \caption{The prefill latency and accuracy on HelloSwag datasets among different quantization algorithms atop Xiaomi 14 using Qualcomm QNN framework.}
    \label{fig:sec-bg-quantization-inference-accuracy-speed}
\end{figure}

%% file: figure-bg-llm-inference.tex
\begin{figure}[t]
    \centering
    \includegraphics[width=0.4\textwidth]{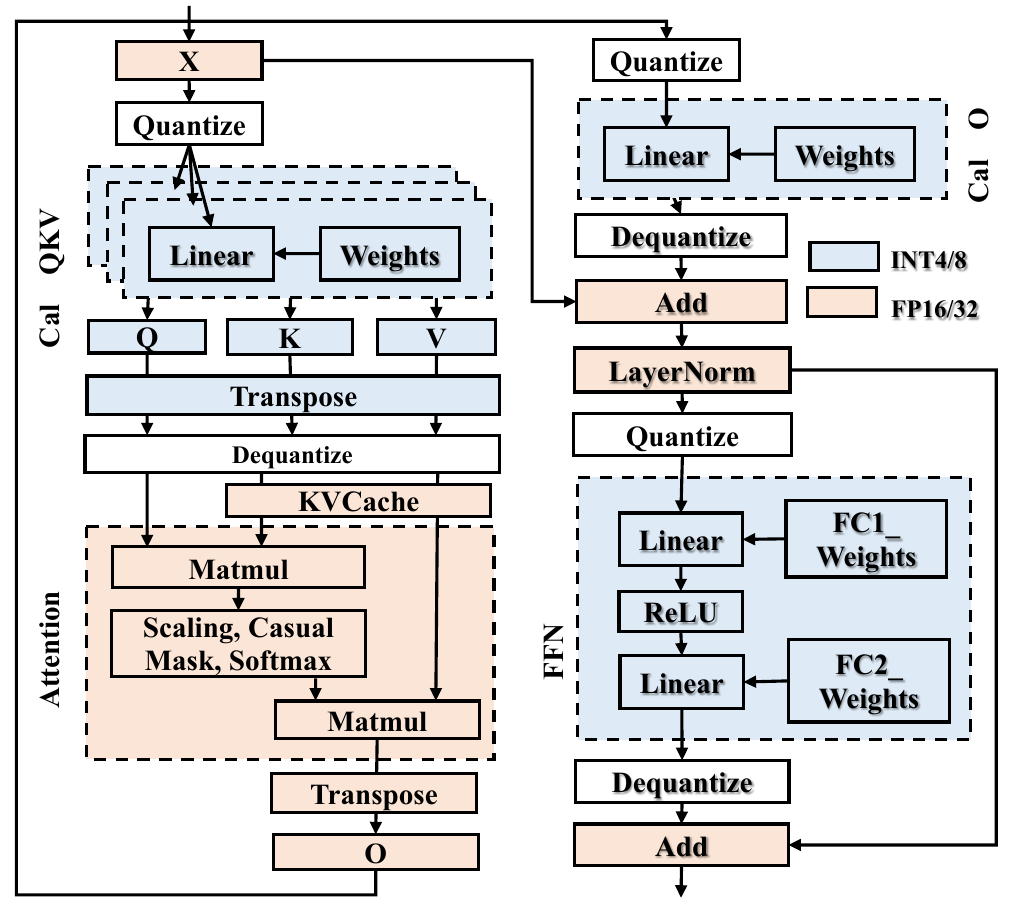}
    \caption{The workflow of quantized on-device LLM inference. Operations shown in blue are computed using INT4/INT8 formats, while those in orange are computed using float data formats.}
    \label{fig:sec-bg-llm-inference}
\end{figure}

%% file: table-bg-module-format.tex
\begin{table}[t]
\centering
\caption{The operator data formats in the state-of-the-art quantization inference approaches. Each module in the first row is illustrated in Figure~\ref{fig:sec-bg-llm-inference}.}
\label{tab:sec-bg-module-format}
\resizebox{0.48\textwidth}{!}{%
\begin{tabular}{llllllll}
\hline
\textbf{Quantization} & \textbf{Type} & \textbf{Acc.} & \textbf{Cal QKV} & \textbf{Atten.} & \textbf{Cal O} & \textbf{Norm.} & \textbf{FFN} \\ \hline
K-Quant~\cite{llama-cpp}               & Per-Group     & Low            & INT8             & FP16               & INT8           & FP16           & INT8         \\ \hline
GPTQ~\cite{frantar2022gptq}                  & Per-Group     & High           & FP16             & FP16               & FP16           & FP16           & FP16         \\ \hline
AWQ~\cite{lin2023awq}                   & Per-Group     & High           & FP16             & FP16               & FP16           & FP16           & FP16         \\ \hline
SmoothQuant~\cite{xiao2023smoothquant}           & Per-tensor    & Low            & INT8             & FP16               & INT8           & FP16           & INT8         \\ \hline
\multicolumn{8}{l}{"Atten.":Attention; "Norm.": Normalization.}
\end{tabular}%
}
\end{table}

%% file: sec-design.tex
\section{\sys Design}

\begin{figure}[t]
    \centering
    \includegraphics[width=0.48\textwidth]{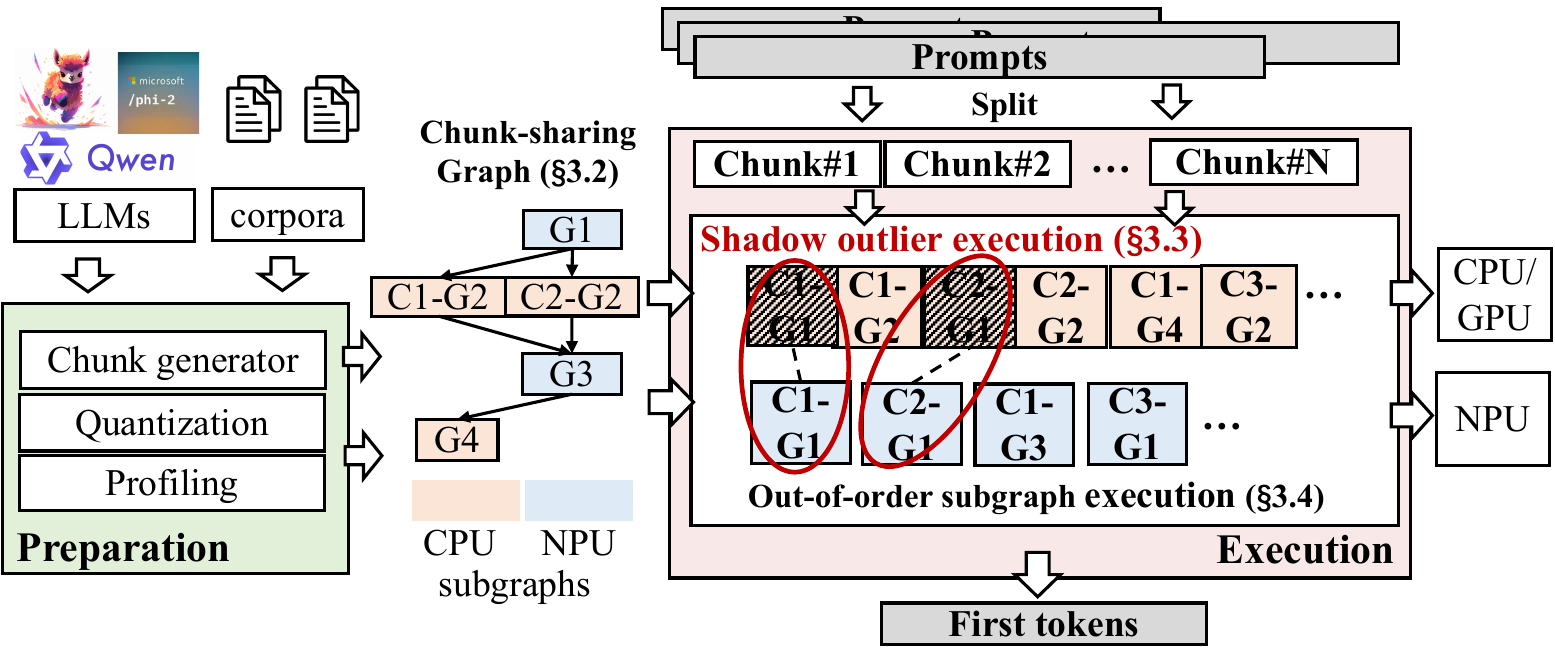}
    \caption{The workflow of \sys.}
    \label{fig:sec-design-workflow}
\end{figure}


\subsection{Overview of \sys}

\textbf{Design goal.}
\sys aims to reduce prefill latency and energy consumption for mobile-sized LLMs through on-device NPU offloading. It supports various mobile-sized LLMs on devices and can be integrated as part of LLM-as-a-System-Service in mobile OS or mobile application services~\cite{yin2024llm,yuan2024mobile}.

\noindent \textbf{Workflow.}
Figure~\ref{fig:sec-design-workflow} illustrates the workflow of \sys.
The key idea of \sys is to maximize its execution on mobile NPU for integer operation acceleration; while keep necessary floating point operations on CPU/GPU to not compromise accuracy.
To enable more efficient NPU offloading, \sys re-constructs the prompt and model in following ways:
(1) At \textit{prompt level:} variable-length prompt is reduced into multiple fixed-sized chunks with data dependency preserved;
(2) At \textit{block level:} Transformer block is scheduled into CPU/GPU and NPU based on their hardware infinity and accuracy sensitivity;
(3) At \textit{tensor level:} important outliers are identified and extracted to run on CPU/GPU.

\noindent $\bullet$ \textit{Preparation stage.}
\sys first uses an enhanced per-tensor quantization algorithm to quantize LLMs into W8A8 format.
The quantization algorithm differs from existing ones as it filters out most unimportant activation outliers and extracts the rest of them into independent, lightweight operators that are complementary to the original one.
\sys also generates fixed-length \textit{chunk-sharing graphs} ($\S$\ref{sec:design-chunk}) to efficiently handle variable-length prompts. 

\noindent $\bullet$ \textit{Execution stage.}
When receiving a prompt, \sys divides it into fixed-sized chunks and processes them causally. 
These chunk graphs will be split into subgraphs scheduled onto the CPU/GPU and NPU  according to their data formats for efficient execution. 
To preserve accuracy, certain INT8-based linear layers undergo sparse float outlier shadow execution on the CPU/GPU in parallel to compensate for quantization errors from outliers  ($\S$\ref{sec:design-outlier}).
To enhance execution efficiency, \sys judiciously schedules the chunks in out-of-order manner ($\S$\ref{sec:design-balance-execution}).

\subsection{Chunk-sharing graph execution} \label{sec:design-chunk}
To tackle the dynamic prompt length challenge, an intuitive solution is to set a fixed length compute graph ahead and use padding~\cite{tillet2019triton,olston2017tensorflow,FastTransformer}. 
However, this method lacks flexibility and excessive padding wastes compute resources.

\input{figure-design-chunk-graph}

\noindent  \textbf{Chunk-wise prefill.}
To enhance flexibility and minimize padding for variable-length prompts, we recognize that processing a long prompt in a LLM is equivalent to processing several split sub-prompts, or ``chunks'', causally. 
This is feasible because popular LLMs use a decoder-only architecture, where the result of the $i$-th token depends only on the preceding tokens.
To that end, \sys first pre-builds and pre-optimizes fixed-length chunk-based NPU compute graph at the preparation stage.
During inference, \sys splits long prompts into several chunks and processes them using these pre-built chunk graphs, as illustrated in Figure~\ref{fig:sec-design-chunk-graph}(b).

However, solely using chunk graphs is not scalable, as \sys would need to store numerous distinct chunk graphs in memory, significantly increasing memory overhead. This is because different chunk graphs have attention operators of varying sizes.
For instance, considering a prompt length of 1024 and a chunk length of 32, the QKV dimension sizes of the attention operators for the first chunk are all $32*hds$, while for the last chunk they are $32*hds$, $1024*hds$, and $1024*hds$ respectively, as shown in Figure~\ref{fig:sec-design-chunk-graph}(b).

\textbf{Chunk-sharing graph.}
\sys introduces a \textit{chunk-sharing graph}, shown in Figure~\ref{fig:sec-design-chunk-graph}(c), based on the insight that LLM operators fall into two distinct categories: (1) static operators (in green), such as Linear and LayerNorm, which depend only on the chunk length and can be shared across different chunks; and (2) dynamic operators (in red), such as Attention, which depend on both chunk length and chunk sequence and cannot be shared among different chunks.
Consequently, \sys divides the LLM into several subgraphs based on the shareability of operators. The shared subgraphs are built and optimized once, whereas non-shared subgraphs are constructed individually for different chunks. During the prefill phase, activations from different chunks pass through the same static operator subgraphs while dynamically selecting the appropriate dimension-specific dynamic operators. This method significantly reduces memory overhead and enhances scalability, as most dynamic operators, like Attention, do not contain weights, requiring only activation buffers.

Our experiments show that 120 out of 144 subgraphs can be shared in Qwen1.5-1.8B models, reducing memory consumption by up to 75\% (7.2GB) for a prompt length as 1024 and a chunk length as 256.


\input{figure-design-scale-chunk-prefill}




We also conducted extensive experiments on selecting a proper chunk length.
The results of two popular LLMs (Qwen1.5-1.8B and Gemma-2B) on Xiaomi 14 device is illustrated in Figure~\ref{fig:sec-design-scale-chunk-prefill}.
Based on the observations, \sys empirically chooses a chunk length of 256 for Xiaomi 14 device, which effectively utilizes the capabilities of mobile NPUs while reducing intra-chunk padding.
In practice, such profiling needs to be performed across different NPUs.

\input{figure-design-outlier-compensate}

\subsection{Shadow outlier execution} \label{sec:design-outlier}







To enable NPU-friendly, per-tensor activation quantization without compromising LLM accuracy, \sys adopts a novel approach termed \textit{shadow outlier execution}.
As shown in Figure~\ref{fig:sec-design-outlier-compensate}, \sys extracts the activation channels with outliers at runtime into a more compact tensor, executes it on CPU, and merges it back to the outcome of original operator on NPU.
This procedure can be formulated as follows:
\begin{equation}
\begin{small} 
    \begin{aligned}
        \frac{x}{s} \odot w =& \left\{\min \left[\max  \left(\frac{x}{s}, -127 \right), 128\right]+\lfloor \frac{x}{s} / 128 \rfloor \times 128 \right\} \odot w \\
                =& \min \left[\max  \left(\frac{x}{s}, -127\right), 128\right] \odot w  \quad \quad \textbf{on NPU} \\
                + & \; extract\left( \lfloor\frac{x}{s} / 128\rfloor \times 128 \right) \odot w \quad \quad \; \textbf{on CPU}
    \end{aligned}
\end{small}
\end{equation}
where $x$, $w$, $s$, $\odot$, and $extract$ represent the original float activation, INT8 weights, the quantization scale factor, the MatMul operation, and the function of extracting activation outliers into a more compact tensor, respectively.
Specifically, the MatMul $\frac{x}{s} \odot w$ can be equivalently divided into the sum of two parts according to the associative law:
(1) \textit{Mobile NPU for MatMul within the scale.} \sys first quantizes and rounds $x$ to the range of -127 to 128 based on the scale factor $s$. It then obtains intermediate results by performing a standard W8A8 per-tensor MatMul with weights $w$.
(2) \textit{Mobile CPUs/GPUs for MatMul beyond the scale.} \sys calculates the partial values exceeding $s$. Since these outliers are rare, \sys extracts these values from the tensor, compresses them into a dense tensor, and performs a MatMul with weights $w$.


Since outliers are very sparse (around 5--15 channels, accounting for only 0.1\%--0.3\% of total channels, as shown in Figure~\ref{fig:sec-design-figure-inference-outlier-states}), the shadow execution on CPU is much faster than the execution of original tensor on NPU, and its execution time can be totally hidden by overlapping.
To further minimize the overhead of this extra process, \sys determines an outlier threshold (i.e., $s$ in Equation 1) by profiling a large corpora at offline, thereby can identify the outliers by simply comparing the activation numbers to this threshold.
The design of shadow outlier execution is compatible with any per-tensor quantization algorithms, and the current prototype of \sys is based on the simple max-min symmetry quantization~\cite{jacob2018quantization}.

While the shadow outlier execution seems to have well balanced the NPU affinity and LLM accuracy, two more crucial issues need to be addressed to make it practical.
First, though mobile SoC uses a unified memory chip for heterogeneous processors, they use separated memory space.
To enable shadow execution of activation outliers, \sys has to keep another copy of each MatMul weights on CPU's memory space.
This increases the memory footprint by almost 2 times.
Second, while the execution of outlier is fast even on CPU, the synchronization of the reduced sum between CPU and NPU still takes non-trivial overhead, e.g.,
29.7\% end-to-end latency and 20.1\% energy consumption on Qwen1.5-1.8B.

\input{figure-design-outlier-stats}
\input{figure-design-outlier-hot}

\textbf{Most outliers tend to appear in a small set of channel positions.}
We observe that, while outliers appear in a wide range of channel positions during processing a long prompt (e.g., 78\%), such appearance is highly skewed -- few channels dominate the appearance of outliers. 
For instance, as shown in Figure~\ref{fig:sec-design-figure-inference-outlier-hot}, less than 3\% of channels contribute to the majority of outliers (over 80\%) across various inputs on Qwen1.5-1.8B and LlaMA-2-7B models.
Therefore, for shadow outlier execution, \sys only keeps the tensor weights that need to be used by those ``hot channels'' in CPU memory space, and retrieve the rest of them from disk if outliers on those positions are extracted (which is rare) at runtime.
Note that the weights retrieval can be also overlapped with the NPU execution of the original MatMul.
This approach reduces the memory overhead of shadow execution by 34.3\% with negligible latency.

\textbf{Most outliers can be pruned without impacts on accuracy.}
Surprisingly, we observe that the activation outliers on most MatMul operators are unimportant to the LLM accuracy and can be simply removed.
Here, the importance of an outlier is measured as the ratio between the largest outlier and the quantization scale ($s$ in Equation 1).
The larger ratio indicates a more dispersed activation distribution, resulting in more significant quantization errors.
\sys profiles these outlier importance using large corpora data at offline stage, (Figure~\ref{fig:sec-design-figure-inference-outlier-pruning}), and prunes most of unimportant layers' outliers.
Typically, we observed layers near the inputs and outputs have a higher importance.
This is because layers close to inputs are easily influenced by the tokens disparity, exhibiting greater fluctuations, while layers approaching to outputs easily accumulate the errors from the shallow layers.
Based on the observation, \sys prunes the outliers of top 85\% most unimportant layers through offline profiling, so that the CPU-NPU synchronization is eliminated.

\input{figure-design-outlier-pruning}


\subsection{Out-of-order subgraph execution} \label{sec:design-balance-execution}
As elaborated in ($\S$\ref{sec-bg-observations}), LLM quantization algorithms cannot fully eliminate floating point operations, \sys thereby divides its execution flow into NPU and CPU/GPU collaboratively.
Typically, \texttt{LayerNorm}, \texttt{Attention}, as well as the shadow outlier computation are placed on the CPU/GPU; while the other linear layers are processed on the NPU.
However, we found simply overlapping their execution is inefficient, resulting in large execution bubbles (37\% bubble rate in critical path), as illustrated in Figure~\ref{fig:sec-design-figure-graph-scheduling}(a).

\noindent \textbf{Out-of-order execution.}
To reduce these execution bubbles, \sys is guided by a key insight that, after being partitioned at both chunk and subgraphs levels, the LLM subgraphs can be scheduled in an out-of-order manner.
More specifically, any input-ready subgraph can be executed without strictly following the chunk sequence.
For instance, the first subgraph of the third chunk (C3-Graph1) can be executed during the bubble period when C2-Graph1 finishes.

\input{figure-desing-chunk-graph-scheduling}

To preserve correctness, \sys considers two types of dependency:
(1) \textit{Cross-chunk dependency.} Operators like Attention rely on data from previous chunks. This means the $i$-th chunk $j$-th subgraph $G_{i,j}$ depends on the $j-1$-th subgraph of the $0, 1, \dots, i-1$ chunks: 
\begin{align}
    G_{i,j} \leftarrow {G_{0,j-1}, G_{1,j-1}, \dots, G_{i,j-1}} 
\end{align}
(2) \textit{Intra-chunk dependency.} Operators like LayerNorm, Linear, and Quantize rely only on previous subgraphs within the same chunk. Therefore, the $i$-th chunk's $j$-th subgraph $G_{i,j}$ depends on the $j-1$-th subgraph of the same chunk:
\begin{align}
    G_{i,j} \leftarrow {G_{i,j-1}} 
\end{align}
As mobile processors are weak at parallelism and preemption~\cite{yi2020heimdall,han2024pantheon,xu2022mandheling}, to ensure efficiency, a processor is capable of executing only one subgraph at any given time.:
\begin{align}
    \sum_{i=0}^N \sum_{j=0}^M P_{i,j,t} = 1, \forall t
\end{align}
where $P_{i, j, t}=1$ indicates that subgraph $G_{i,j}$ is running on processor $P$ at time $t$, and $N$ and $M$ represent the maximum number of chunks and subgraphs, respectively.
\sys aims to find an execution order minimizing the total execution time of all subgraphs under these constraints.
Unfortunately, this scheduling problem can be reduced to a classical NP-Hard  Traveling Salesman Problem~\cite{hoffman2013traveling}.
Moreover, because the number of chunks varies with user prompts, an optimal scheduling strategy cannot be generated offline.

Instead, \sys utilizes an online heuristic algorithm. The key idea is to focus not on the execution time of the subgraph $g$, but on how executing $g$ contributes to reducing NPU stalls, motivated by the observation that during the prefill phase, NPU execution time often dominates inference latency, being the critical path. For instance, with a prompt length of 256 using the Qwen1.5-1.8B model, NPU execution takes 315ms, about twice that of the CPU.

Specifically, we define a subgraph $g$'s contribution to reduce NPU stalls as follows: If subgraph $g$ is to be executed on the CPU/GPU, let $S$ be the set of new subgraphs that can be executed after $g$ is completed. $S$ will be executed on the NPU. A longer execution time of $S$ is beneficial for reducing NPU stalls. Thus, $g$'s contribution is defined as the total execution time of $S$. Conversely, if $g$ is executed on the NPU, a shorter execution time of $S$ is beneficial, with the negative value of $S$'s execution time as $g$'s contribution, formulated as:
\begin{align}
    C = \begin{array}{l} 
  \left\{\begin{matrix} 
  \sum T_i, \forall i \in S \text{  if $g$ is on the CPU/GPU} \\ 
  -\sum T_i, \forall i \in S \text{  if $g$ is on the NPU} \\ 
    \end{matrix}\right.    
    \end{array} 
\end{align}
where $T$ is the subgraph execution time.
\sys always chooses the subgraph with the largest $C$, meaning the subgraph $g$ with $S$ having the longest execution time on the NPU or the shortest execution time on the CPU/GPU.

In a nut shell, \sys profiles all the subgraph execution time and their dependency offline at the preparation stage. During the prefill stage, it calculates all the pending subgraphs $C$ value and selects one with maximum $C$ to run, with microsecond-level performance overhead.

%% file: figure-design-chunk-graph.tex
\begin{figure}[t]
    \centering
    \includegraphics[width=0.48\textwidth]{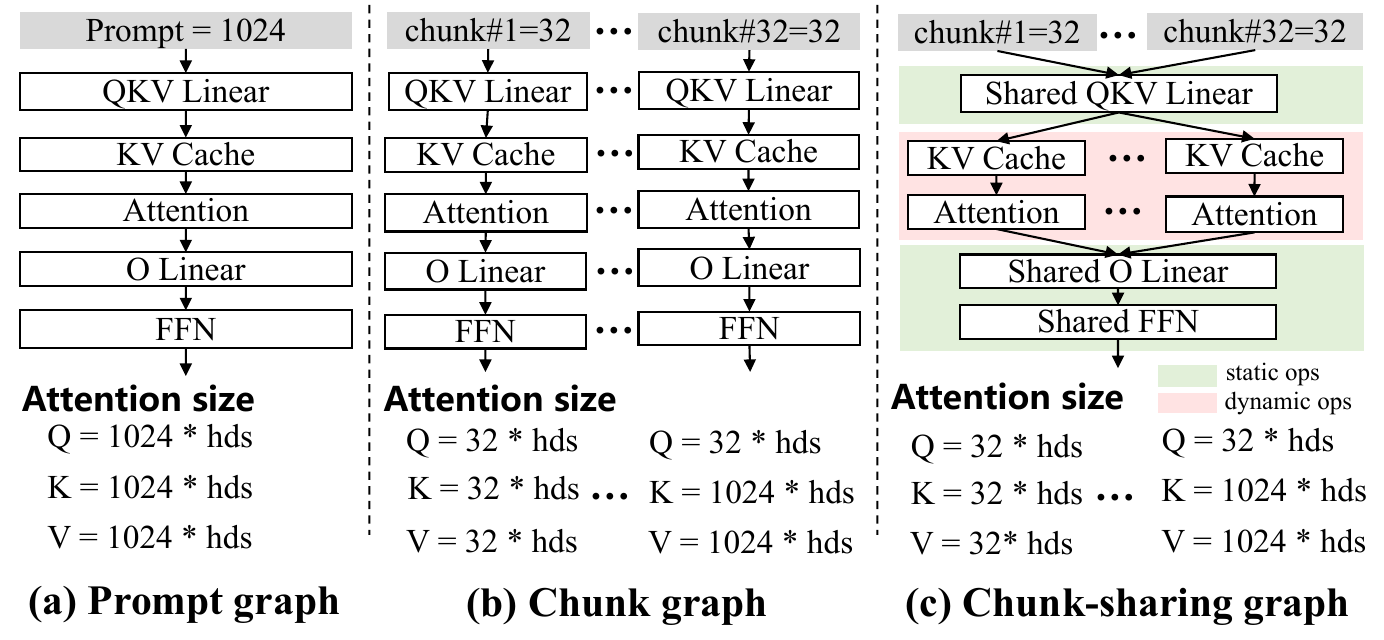}
    \caption{The illustration of prompt graph, chunk graph and chunk-sharing graph. The chunk length is 32.}
    \label{fig:sec-design-chunk-graph}
\end{figure}

%% file: figure-design-scale-chunk-prefill.tex
\begin{figure}[t]
    \centering
    \includegraphics[width=0.48\textwidth]{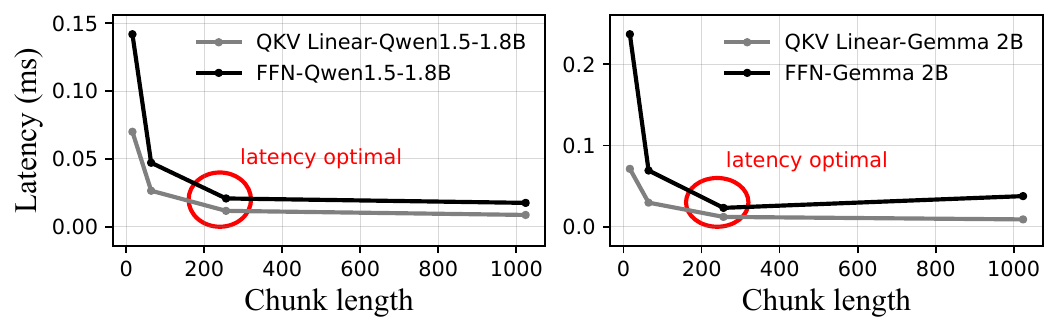}
    \caption{The per-token latency of QKV Linears and FFN under different chunk lengths.}
    \label{fig:sec-design-scale-chunk-prefill}
\end{figure}

%% file: figure-design-outlier-compensate.tex
\begin{figure}[t]
    \centering
    \includegraphics[width=0.45\textwidth]{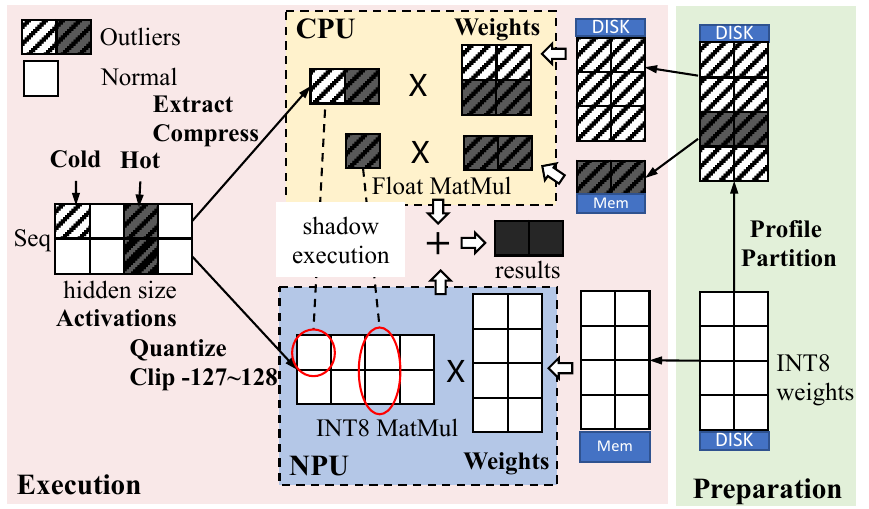}
    \caption{The workflow of shadow outlier execution.}
    \label{fig:sec-design-outlier-compensate}
\end{figure}

%% file: figure-design-outlier-stats.tex
\begin{figure}
    \centering
    \includegraphics[width=0.48\textwidth]{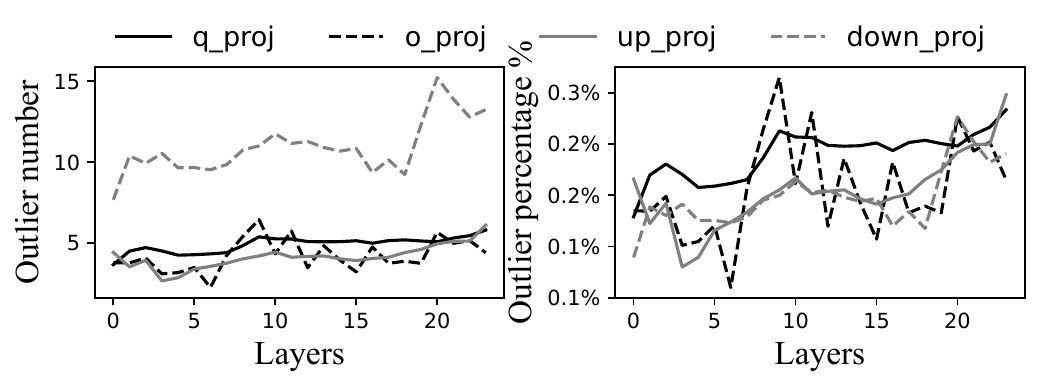}
    \caption{The average number and percentage of outlier channels per layer on Qwen1.5-1.8B model using the wikitext dataset under 2048 inference. \textbf{Less than 0.3\% channels have outliers during one inference.}}
    \label{fig:sec-design-figure-inference-outlier-states}
\end{figure}

%% file: figure-design-outlier-hot.tex
\begin{figure}
    \centering
    \includegraphics[width=0.48\textwidth]{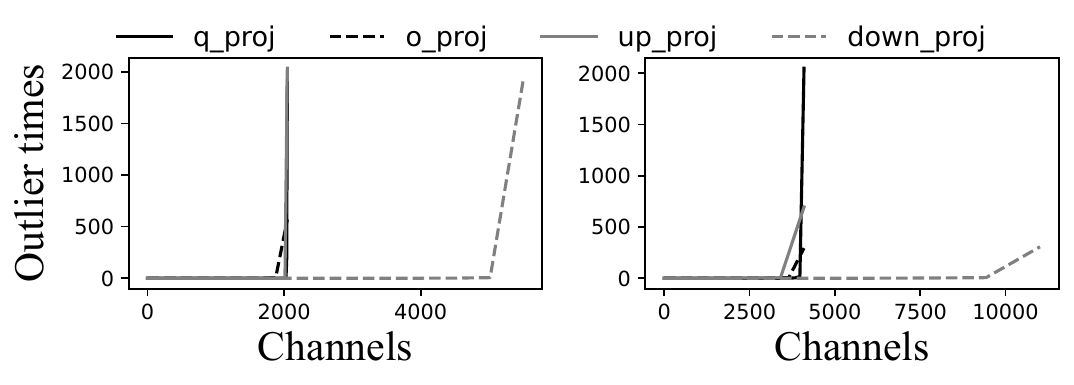}
    \caption{The outlier times per channel on Qwen1.5-1.8B model using the wikitext dataset under 2048 inference. \textbf{Less than 3\% channels contribute over 80\% outliers.}}
    \label{fig:sec-design-figure-inference-outlier-hot}
\end{figure}

%% file: figure-design-outlier-pruning.tex
\begin{figure}
    \centering
    \includegraphics[width=0.48\textwidth]{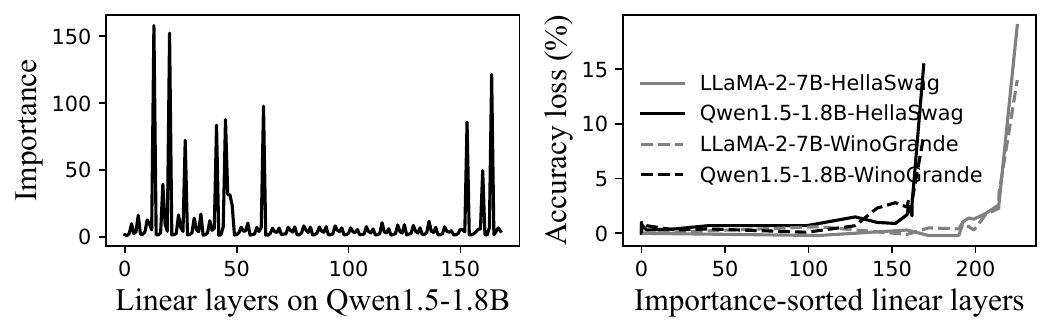}
    \caption{Left: Outlier importance of different layers in Qwen1.5-1.8B model. Right: Relationship between accuracy and pruned layers on HellaSwag and Winograde datasets.}
    \label{fig:sec-design-figure-inference-outlier-pruning}
\end{figure}

%% file: figure-desing-chunk-graph-scheduling.tex
\begin{figure}[t]
    \centering
    \includegraphics[width=0.45\textwidth]{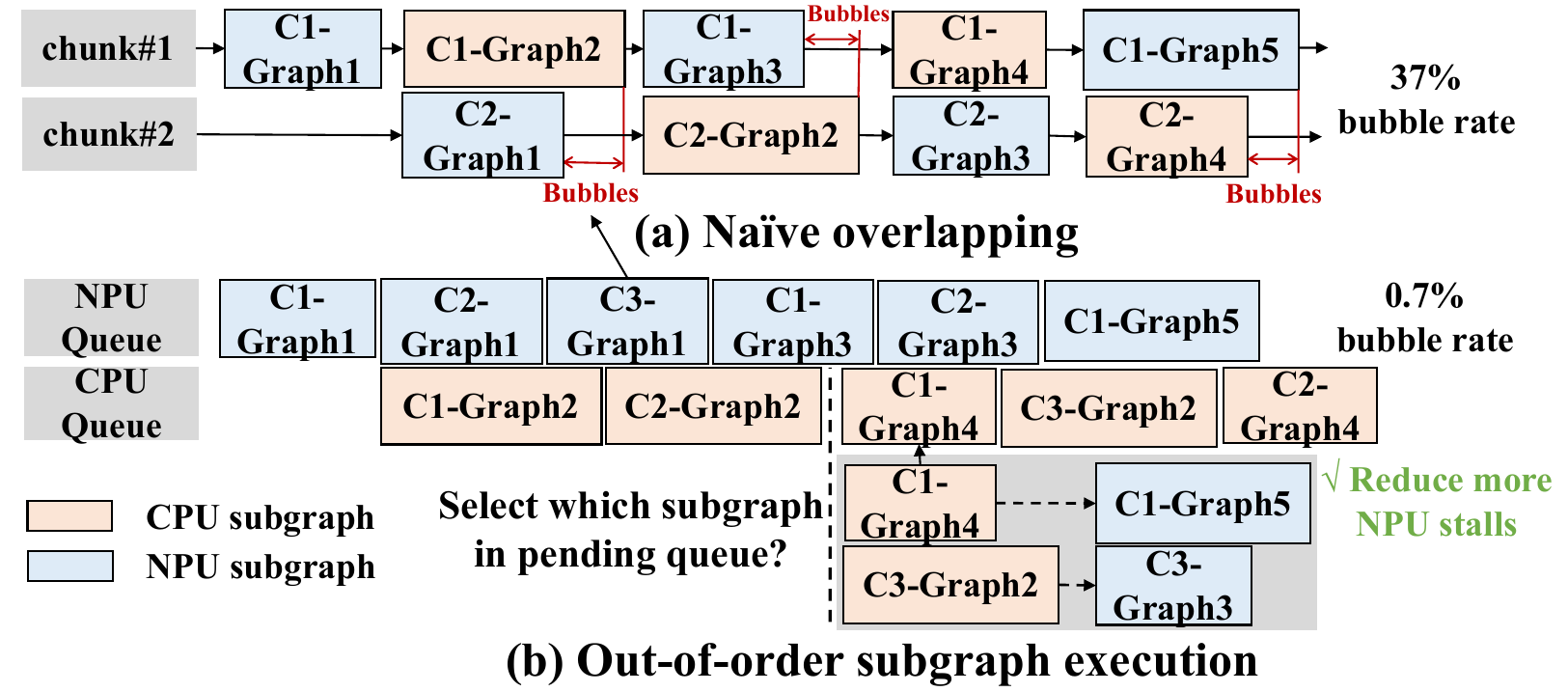}
    \caption{Out-of-order  subgraph execution.}
    \label{fig:sec-design-figure-graph-scheduling}
\end{figure}

%% file: sec-eval.tex
\input{figure-eval-seq-prefill}

\section{Implementation and Evaluation}
We have fully implemented \sys for Qualcomm Hexagon NPUs, comprising 10K lines of code in C/C++ and assembly language.
We choose Qualcomm SoCs as the target platform for its popularity on mobile devices and powerful NPU capacity.
Qualcomm Hexagon is also the only mobile NPU with an open instruction set architecture.
\sys is built on the MLLM~\cite{mllm}, one state-of-the-art mobile LLM engines, and QNN framework~\cite{QNN}, the Qualcomm Neural Processing SDK.
It supports standard LLM formats exported from Hugging Face~\cite{Hugging-Face}.
To facilitate LLM execution, we implemented specific operators like KVCache, SiLU, RMSNorm, ROPE, and etc, in addition to what have been supported by QNN.
To reduce context switching overhead between CPUs/GPUs and NPUs, \sys leverages shared buffers to synchronize intermediate results from different processors.
For end-to-end inference, \sys is compatible with any decoding engine and utilizes the MLLM CPU backend for decoding stage as easy implementation, with a default chunk length of 256.
The default pruning rate for outlier layers is 85\%.
\revision{\sys does not consider resource detection and contention in the current implementation.}

The prototype further incorporates two optimizations.
(1)
Our extensive experiments show that mobile NPUs ofte favor tensor sizes (e.g., equal ``height'' and ``width'') in CNN architectures.
For example, a linear layer with weights of 2048$\times$2048 produces the same results for inputs of 1024×1×2048 and 32$\times$32$\times$2048, but using 32$\times$32$\times$2048 reduces execution latency by 1.62$\times$.
Therefore, \sys profiles all possible equivalent shapes for linear layers during the preparation stage and selects the most efficient one.
(2)
Mobile NPUs typically access limited memory regions (e.g., 4GB for Hexagon NPU), which can be smaller than the size of LLM weights. To maximize prefill acceleration within this limited memory, \sys prioritizes executing computationally intensive tasks, such as FFN, on the NPU to enhance efficiency.

\subsection{Experiment setups}
\textbf{Hardware setup.}
We test \sys on two smartphones with different Qualcomm SoCs: Redmi K70 Pro (Snapdragon 8gen3, 24GB memory) and Redmi K60 Pro (Snapdragon 8gen2, 16GB memory).
All devices run Android OS 13.

\noindent \textbf{Models and datasets.}
We conducted tests using a variety of typical mobile-sized LLMs: Qwen1.5-1.8B~\cite{qwen2}, Gemma-2B~\cite{gemma2b}, Phi2-2.7B~\cite{phi2-2.7}, LLaMA2-Chat-7B~\cite{Llama-2-7b}, and Mistral-7B~\cite{Mistral-7B}. To evaluate \sys's quantization accuracy, we employed widely recognized LLM benchmarks, including LAMBADA~\cite{lambada}, HellaSwag~\cite{zellers2019hellaswag}, WinoGrande~\cite{ai2:winogrande}, OpenBookQA~\cite{OpenBookQA2018} and MMLU~\cite{hendryckstest2021}. For inference speed experiments, we selected retrieval-based datasets from Longbench, 2wikimqa and TriviaQA~\cite{bai2023longbench}, for simulating context-aware generate tasks like automated email reply. Additionally, we assessed \sys in screen question-answering and mapping instruction to UI action scenarios using DroidTask datasets~\cite{LlamaTouch} to simulate agent-based UI automation tasks.

\noindent \textbf{Baselines.}
We mainly compare \sys with 5 baselines, including 3 widely used mobile LLM engines (\texttt{TFLite}~\cite{tflite}, \texttt{MNN}~\cite{mnn}, and \texttt{llama.cpp}~\cite{llama-cpp}).
Those engines support only mobile CPU and GPU;
2 advanced baselines are \texttt{MLC-LLM}~\cite{mlc-llm}, an LLM compiler for on-device GPUs, and \texttt{PowerInfer-v2}, which also utilizes mobile NPUs to accelerate prefilling~\cite{xue2024powerinfer}. 
Since PowerInfer-v2 is not open-sourced, we use the reported data from its paper.
To be noted, those baselines often support only a subset of 5 LLMs we evaluated.

\noindent \textbf{Metrics.}
We mainly measure LLM inference accuracy, prefill latency, prefill energy consumption, prefill memory consumption and end-to-end inference latency.
The energy consumption is obtained through /sys /class/power\_supply in Android OS
by profiling every 100ms.
All experiments are repeated three times and we report the average
numbers.

\input{figure-eval-end-to-end}

\subsection{Prefill performance.}
We evaluate the \sys's prefill performance (speed and energy consumption) at prompt lengths of 64, 256 and 1024 tokens on two devices, as shwon in  Figure~\ref{fig:sec-eval-sequence-prefill} and \ref{fig:sec-eval-sequence-prefill-energy}.
Despite outlier variations across datasets, the overall impact on prefill performance is minimal, so we report results from the LongBench dataset.
The results show that \textit{\sys consistently outperforms all baselines across both metrics, with benefits becoming more pronounced as prompt length increases.}

\noindent \textbf{Prefill speed.}
For prompt length of 1024 tokens, \sys can reduce prefill latency by 18.17--38.4$\times$, 7.3$\times$, 32.5--43.6$\times$, and 1.27--2.34$\times$ on Redmi K70 Pro compared with llama.cpp-CPU, MNN-CPU, MLC-GPU, TFLite-GPU, respectively.
On the Redmi K60 Pro, these improvements are 21.3--41.3$\times$, 7.43$\times$, 37.2--69.3$\times$, and 1.3--2.6$\times$, respectively.
These speedups are due to \sys's use of three innovative techniques that fully leverage mobile NPUs, including \textit{shadow outlier execution}, high-efficiency per-tensor MatMul, and  \textit{out-of-order subgraph execution}.
Compared with PowerInfer-V2-NPU, a baseline also using NPU for prefilling, \sys can achieve 3.28--5.32$\times$ and 3.4--5.6$\times$ speedup on two devices, respectively, by employing NPU-friendly INT8 linear computation and fine-grained subgraph scheduling ($\S$\ref{sec:design-balance-execution}).

For prompt length of 64 tokens, the prefill speed of  \sys is 14.86--7.10$\times$, 1.69$\times$, 10.91--17.32$\times$, 1.48$\times$, and 1.81--2.51$\times$ faster than llama.cpp-CPU, MNN-CPU, MLC-GPU, TFLite-GPU, and PowerInfer-V2-NPU respectively, with speedups averaging 10.5$\times$, 4.31$\times$, 2.68$\times$, 1.02$\times$, and 1.96$\times$ lower than those for 1024-token prompts.
This is because a shorter prompt can lead to a padding problem and limit \sys's out-of-order execution scheduling efficiency.

\noindent \textbf{Prefill energy consumption.}
Energy consumption was evaluated on the Redmi K60 Pro, the only rootable device.
PowerInfer-V2 was excluded due to the lack of energy consumption data and open-source code. 
For 1024-token prompts, \sys reduces energy consumption by 35.63--59.52$\times$, 35.21-- 59.25$\times$, and 1.85--4.32$\times$ compared to llama.cpp-CPU, MLC-GPU, and TFLite-GPU, respectively. For 64-token prompts, the savings are 10.38--14.12$\times$, 10.38--17.79$\times$, and 3.22--3.67$\times$, respectively. These savings are due to the high energy efficiency of mobile NPUs and \sys's three novel techniques for maximizing NPU performance.
\revision{
Typically, during the LLM prefill stage, all CPU cores are fully utilized, consuming the highest power; NPUs operate at 500-750 MHz, consuming the least power.}






\subsection{End-to-end performance}
\revision{We evaluate the real-world performance of \sys against baseline systems using three workloads: UI automation on DroidTask datasets, context-aware automated email replies on LongBench datasets, chat summaries on Persona-Chat datasets. 
Table~\ref{tab:sec-eval-end-to-end} presents the prompt and output lengths of the three datasets, along with their corresponding prefill, decode, and end-to-end inference latency results. End-to-end speedups in Table~\ref{tab:sec-eval-end-to-end} are the geometric mean of each sample.}
Our key observation is that \textbf{\sys always achieves the lowest inference latency across all three datasets.} 

For LongBench datasets, \sys shows significant speed improvements: 23.0--46.2$\times$ over llama.cpp-CPU, 16.5--36.4$\times$ over MLC-LLM-GPU, 4.08--4.19$\times$ over MNN-CPU, 3.51--3.73$\times$ over PowerInfer-V2-NPU, and 1.27--2.03$\times$ over TFLite-GPU.
This impressive performance is primarily due to \sys's superior efficiency during the prefill stage. The speedup against TFLite-GPU is lower since \sys currently relies on a CPU backend for decoding with no optimization, while TFLite utilizes GPU. Notably, \sys is compatible with any decoding engine, which means
once TFLite is open-sourced, \sys can integrate it as the decoding backend, potentially enhancing performance further.

For the DroidTask datasets, \sys reduces end-to-end inference latency by 7.9--12.5$\times$ compared to llama.cpp-CPU, 15.0--32.8$\times$ compared to MLC-LLM-GPU, 2.38--2.45$\times$ compared to MNN-CPU, 2.27--2.44$\times$ compared to PowerInfer-V2-NPU, and 1.35--2.38$\times$ compared to TFLite-GPU. The performance gains are slightly smaller for DroidTask datasets due to shorter prompts in UI automation versus email writing.

\revision{
For the Persona-Chat datasets, \sys reduces end-to-end inference latency by an average of 3.5$\times$ compared to llama.cpp-CPU, 10.5$\times$ compared to MLC-LLM-GPU, 3.1$\times$ compared to MNN-CPU, 1.1 times compared to PowerInfer-V2-NPU, and 1.1$\times$ compared to TFLite-GPU. The primary reason for the reduced performance improvement relative to other datasets is the larger number of output tokens generated by this dataset. Since \sys currently supports only CPU-based decoding,  this slower decoding backend results in increased end-to-end latency. This issue can be mitigated by utilizing a mobile GPU as the decoding backend (see Section~\ref{sec:eval-gpu} for details).}

\subsection{Inference accuracy}

\textbf{Overall accuracy.}
We investigate the inference accuracy of \sys on 5 LLM benchmarks: LAMBADA~\cite{lambada}, MMLU~\cite{hendryckstest2021}, WinoGrande~\cite{ai2:winogrande}, OpenBookQA~\cite{OpenBookQA2018} and HellaSwag~\cite{zellers2019hellaswag}.
For comparison, we evaluated 4 alternatives:
FP16 (non-quantization), K-Quant~\cite{llama-cpp} (used in \texttt{llama.cpp}), SmoothQu-ant (state-of-the-art per-tensor method)~\cite{xiao2023smoothquant}, and LLM.Int8() (state-of-the-art float outlier handling method)~\cite{dettmers2022gpt3}.
\textbf{\sys achieves negligible accuracy loss, and significantly outperforms the other quantization algorithms}, as shown in Table~\ref{tab:sec-eval-accuracy}.

Specifically, \sys is, on average, only 1\% less accurate than FP16, and it shows an accuracy improvement of up to 32.9\% over SmoothQuant and up to 70.9\% over K-Quant.
This improvement over SmoothQuant, which uses static profiling to smooth outliers to normal values, is due to \sys's dynamic handling of outlier positions with CPU float precision. \sys addresses outliers at the element level, providing higher precision than K-Quant that uses group-level quantization scales.
Furthermore, \sys achieves comparable accuracy (0.1\% average loss) to LLM.Int8(), as both handle outliers with float precision. But 
\sys better utilizes NPU-specific computational features, maintaining high accuracy and NPU efficiency.

\input{table-eval-accuracy}

\input{figure-trade-off-speed-accuracy}

\noindent \textbf{Trade-offs between accuracy and performance.}
\revision{\sys balances generation accuracy and speed by adjusting outlier pruning rates, \textbf{with higher pruning rates leading to faster generation but lower quality.}
To comprehensively show how outlier pruning rates influence generate accuracy and speed, we conducted experiments using Qwen1.5-1.8B and Gemma-2B on the Redmi K70 Pro with the HellaSwag~\cite{zellers2019hellaswag} and LAMBADA~\cite{lambada} datasets. The results are shown in Figure~\ref{fig:sec-eval-trade-off-accuracy-speed}.}

\revision{
When no outliers are pruned (pruning rate = 0\%), \sys achieves the highest accuracy, albeit with the slowest generation speed. Specifically, the Qwen1.5-1.8B model can attain an accuracy of 71.3\% on the Lambada dataset, with a corresponding generation speed of 156 tokens per second. Conversely, the Gemma-2B model demonstrates an accuracy of 59.6\% and a generation speed of 102 tokens per second. 
Upon pruning 80\% of outliers, the Qwen1.5-1.8B model's accuracy decreases to 42.2\%, while its generation speed increases to 544 tokens per second. Similarly, the Gemma-2B model's accuracy declines to 57.4\%, yet its generation speed rises to 285 tokens per second. When all outliers are pruned, generation accuracy reaches its lowest point, while generation speed is at its highest. Specifically, the Qwen1.5-1.8B model demonstrates an accuracy of merely 8.1\% with a generation speed of 596 tokens/s; the Gemma-2B model reports an accuracy of 41.9\% alongside a generation speed of 345 tokens/s.
}

\subsection{Memory consumption}
We compare \sys with INT8 weight baselines, as mobile NPUs only support INT8 weight computations.  Memory consumption results on the Redmi K60 Pro, using a 512-token prompt, are presented in Figure~\ref{fig:sec-eval-memory}. \sys consumes up to 1.32 $\times$ more memory than llama.cpp and TFLite.
The overhead is due to the MLLM and QNN frameworks, which allocate independent activation buffers for each operator to enhance speed. The tiny additional memory overhead introduced by \sys is its \textit{$\S$\ref{sec:design-outlier} shadow outlier execution} technique (in black), which loads tiny float weights into memory, accounting for only 0.6\%--1\% of the total memory.

\input{figure-eval-memory}

\input{figure-eval-gpu}

\subsection{GPU-NPU coordination analysis} \label{sec:eval-gpu}
\revision{
\sys is compatible with mobile GPUs; however, we have currently implemented only the CPU-NPU version due to easy implementation. To illustrate the potential advantages of employing a mobile GPU, we simulate GPU-NPU co-scheduling, assuming TFLite serves as the GPU backend. The evaluation is conducted using the Gemma-2B model across varying prompt lengths and the Longbench dataset, specifically on the Redmi K70 Pro, as shown in Figure~\ref{fig:sec-eval-gpu}. Our observations indicate that \textbf{while GPU-NPU coordination does not enhance the prefilling speed, it effectively reduces the end-to-end latency.}}

\revision{As illustrated in Figure~\ref{fig:sec-eval-gpu}(a), the prefill speed achieved through GPU-NPU coordination is equivalent to that of CPU-NPU coordination. This phenomenon occurs because, within the design of \sys, the computational demands placed on mobile CPUs and GPUs are significantly lower than those on the NPU (details in Section~\ref{sec:design-balance-execution}). Consequently, the computational latency of mobile CPUs and GPUs can be effectively hidden by the NPU computation, indicating that the choice of a mobile CPU or mobile GPU is not essential. On the other hand, GPU-NPU coordination can decrease end-to-end latency by 80--90 ms compared to CPU-NPU coordination. This improvement is attributed to the faster decoding speed of GPUs relative to CPUs.}

\input{figure-eval-ablation}

\subsection{Ablation study.}
We conduct a comprehensive breakdown analysis of the benefits brought by each of \sys’s techniques using the Qwen1.5-1.8B, Gemma-2B, and LlaMA2-7B models, as shown in Figure~\ref{fig:sec-eval-ablation}.
The leftmost bar represents the prefill speed on the CPU using llama.cpp. The second bar shows a naive implementation on mobile NPUs, followed by the incorporation of our three techniques, with the rightmost bar representing \sys.
The three techniques are represented by \textit{chunk ($\S$\ref{sec:design-chunk})}, \textit{outlier ($\S$\ref{sec:design-outlier})}, and \textit{OOE ($\S$\ref{sec:design-balance-execution})}, respectively.
We observe that \textbf{\sys’s three techniques make a significant contribution to the overall improvement.} 

Firstly, directly offloading LLM prefilling workloads to mobile NPUs results in 2.55--2.68$\times$ delays,  due to the substantial gap between LLMs and mobile NPUs, as detailed in $\S$\ref{sec-bg-observations}.
Additionally, \textit{chunk-sharing graph} improves prefill speed by 1.46--5.09$\times$ by reducing graph building and optimization delays. The Gemma-2B model achieves the highest speedup, as it requires more time to build and optimize.
Furthermore, \textit{shadow outlier execution} reduces prefill latency by 3.91--8.68$\times$,  allowing per-tensor MatMul operations to fully utilize NPUs with minimal CPU/GPU overhead.
Lastly, the \textit{out-of-order subgraph execution} reduces prefill latency by 18\%--44\% by reducing execution bubbles within the NPU.

%% file: figure-eval-seq-prefill.tex
\begin{figure*}
    \centering
     \includegraphics[width=0.98\textwidth]{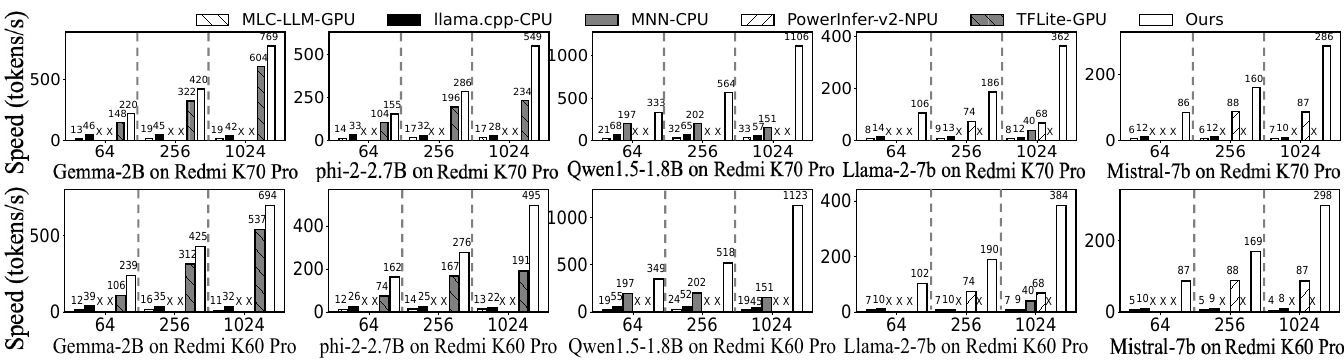}
    \caption{Prefill speed under different prompt lengths on different devices (datasets: Longbench-2wiki-Multi-doc QA).}
    \label{fig:sec-eval-sequence-prefill}
\end{figure*}

\begin{figure*}
    \centering
    \begin{subfigure}[t]{0.98\textwidth}
        \includegraphics[width=\textwidth]{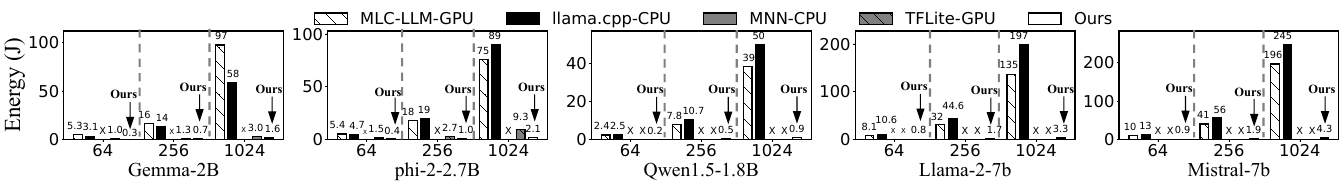}
    \end{subfigure}
        \caption{Energy consumption under different prompt lengths on Redmi K60 Pro (datasets: Longbench-2wiki-Multi-doc QA).}
    \label{fig:sec-eval-sequence-prefill-energy}
\end{figure*}

%% file: figure-eval-end-to-end.tex
\begin{table*}[t]
\caption{\revision{End-to-end latency comparison using real mobile applications execution on Redmi K70 Pro. Each cell is organized in the following format: end-to-end latency (prefill latency, decode latency). Unit: seconds.}}
\resizebox{0.98\textwidth}{!}{%
\begin{tabular}{llrrrrrrr}
\hline
\multicolumn{1}{l|}{\textbf{LLM}}                & \multicolumn{1}{l|}{\textbf{Datasets}}                                                                                                                                        & \textbf{MLC}                                                   & \textbf{LCPP}                                                  & \textbf{MNN}                                                 & \textbf{PI}                                                  & \multicolumn{1}{r}{\textbf{TFLite}}                       & \multicolumn{1}{r|}{\textbf{Ours}}                                              & \textbf{Speedup}                   \\ \hline
\multicolumn{1}{l|}{Qwen1.5-1.8B}                & \multicolumn{1}{l|}{\multirow{5}{*}{\begin{tabular}[c]{@{}l@{}}Longbench: 2wiki-Multi-doc QA \\ prompt length: 1451--1672 tokens\\  output length: 2--4 tokens\end{tabular}}} & \begin{tabular}[c]{@{}r@{}}45.6\\ (45.43, 0.17)\end{tabular}   & \begin{tabular}[c]{@{}r@{}}26.7\\ (26.43, 0.24)\end{tabular}   & \begin{tabular}[c]{@{}r@{}}10.6\\ (10.01, 0.59)\end{tabular} & -                                                            & -                                                          & \multicolumn{1}{r|}{\begin{tabular}[c]{@{}r@{}}1.7\\ (1.49, 0.24)\end{tabular}} & \textbf{6.2-26.8$\times$} \\ \cline{1-1} \cline{3-9} 
\multicolumn{1}{l|}{Gemma-2B}                    & \multicolumn{1}{l|}{}                                                                                                                                                         & \begin{tabular}[c]{@{}r@{}}78.4\\ (78.03, 0.32)\end{tabular}   & \begin{tabular}[c]{@{}r@{}}34.6\\ (34.31, 0.25)\end{tabular}   & -                                                            & -                                                            & \begin{tabular}[c]{@{}r@{}}2.6\\ (2.40, 0.19)\end{tabular} & \multicolumn{1}{r|}{\begin{tabular}[c]{@{}r@{}}1.9\\ (1.68, 0.25)\end{tabular}} & \textbf{1.4-41.3$\times$} \\ \cline{1-1} \cline{3-9} 
\multicolumn{1}{l|}{Phi-2-2.7B}                  & \multicolumn{1}{l|}{}                                                                                                                                                         & \begin{tabular}[c]{@{}r@{}}87\\ (86.70, 0.28)\end{tabular}     & \begin{tabular}[c]{@{}r@{}}53.3\\ (53.00, 0.34)\end{tabular}   & \begin{tabular}[c]{@{}r@{}}13\\ (11.98, 1.05)\end{tabular}   & -                                                            & \begin{tabular}[c]{@{}r@{}}6.3\\ (6.06, 0.28)\end{tabular} & \multicolumn{1}{r|}{\begin{tabular}[c]{@{}r@{}}3.1\\ (2.79, 0.34)\end{tabular}} & \textbf{2.0-28.1$\times$} \\ \cline{1-1} \cline{3-9} 
\multicolumn{1}{l|}{LlaMA-2-7B}                  & \multicolumn{1}{l|}{}                                                                                                                                                         & \begin{tabular}[c]{@{}r@{}}184.7\\ (184.20, 0.45)\end{tabular} & \begin{tabular}[c]{@{}r@{}}146\\ (145.25, 0.70)\end{tabular}   & \begin{tabular}[c]{@{}r@{}}22.4\\ (22.35, 0.05)\end{tabular} & \begin{tabular}[c]{@{}r@{}}19.8\\ (19.04, 0.78)\end{tabular} & -                                                          & \multicolumn{1}{r|}{\begin{tabular}[c]{@{}r@{}}5.3\\ (4.62, 0.70)\end{tabular}} & \textbf{3.7-34.8$\times$} \\ \cline{1-1} \cline{3-9} 
\multicolumn{1}{l|}{Mistral-7b}                  & \multicolumn{1}{l|}{}                                                                                                                                                         & \begin{tabular}[c]{@{}r@{}}254.2\\ (253.45, 0.72)\end{tabular} & \begin{tabular}[c]{@{}r@{}}200.2\\ (199.37, 0.87)\end{tabular} &                                                              & \begin{tabular}[c]{@{}r@{}}20\\ (19.22, 0.78)\end{tabular}   & -                                                          & \multicolumn{1}{r|}{\begin{tabular}[c]{@{}r@{}}5.5\\ (4.61, 0.87)\end{tabular}} & \textbf{3.6-46.2$\times$} \\ \hline
\multicolumn{2}{l|}{\textbf{Geo-mean (speedup)}}                                                                                                                                                                                 & \textbf{34.7$\times$}                                          & \textbf{21.8$\times$}                                          & \textbf{4.8$\times$}                                         & \textbf{3.7$\times$}                                         & \textbf{1.7$\times$}                                       & \multicolumn{1}{r|}{\textbf{-}}                                                 &                           \\ \hline
\multicolumn{1}{l|}{Qwen1.5-1.8B}                & \multicolumn{1}{l|}{\multirow{5}{*}{\begin{tabular}[c]{@{}l@{}}Longbench:  TriviaQA \\ prompt length: 1511--1787 tokens\\ output length: 5--11 tokens)\end{tabular}}}         & \begin{tabular}[c]{@{}r@{}}46\\ (45.68)\end{tabular}           & \begin{tabular}[c]{@{}r@{}}27\\ (26.57, 0.47)\end{tabular}     & \begin{tabular}[c]{@{}r@{}}11.2\\ (10.07, 1.15)\end{tabular} & -                                                            & -                                                          & \multicolumn{1}{r|}{\begin{tabular}[c]{@{}r@{}}2.0\\ (1.49, 0.47)\end{tabular}} & \textbf{5.6-23.0$\times$} \\ \cline{1-1} \cline{3-9} 
\multicolumn{1}{l|}{Gemma-2B}                    & \multicolumn{1}{l|}{}                                                                                                                                                         & \begin{tabular}[c]{@{}r@{}}81.8\\ (81.25, 0.26)\end{tabular}   & \begin{tabular}[c]{@{}r@{}}36.2\\ (35.72, 0.43)\end{tabular}   & -                                                            & -                                                            & \begin{tabular}[c]{@{}r@{}}2.8\\ (2.50, 0.32)\end{tabular} & \multicolumn{1}{r|}{\begin{tabular}[c]{@{}r@{}}2.2\\ (1.75, 0.43)\end{tabular}} & \textbf{1.3-37.2$\times$} \\ \cline{1-1} \cline{3-9} 
\multicolumn{1}{l|}{Phi-2-2.7B}                  & \multicolumn{1}{l|}{}                                                                                                                                                         & \begin{tabular}[c]{@{}r@{}}91.4\\ (90.94, 0.50)\end{tabular}   & \begin{tabular}[c]{@{}r@{}}56.3\\ (55.60, 0.68)\end{tabular}   & \begin{tabular}[c]{@{}r@{}}14.7\\ (12.57, 2.11)\end{tabular} & -                                                            & \begin{tabular}[c]{@{}r@{}}6.8\\ (6.36, 0.40)\end{tabular} & \multicolumn{1}{r|}{\begin{tabular}[c]{@{}r@{}}3.6\\ (2.94, 0.68)\end{tabular}} & \textbf{1.9-25.4$\times$} \\ \cline{1-1} \cline{3-9} 
\multicolumn{1}{l|}{LlaMA-2-7B}                  & \multicolumn{1}{l|}{}                                                                                                                                                         & \begin{tabular}[c]{@{}r@{}}197.3\\ (196.43, 0.83)\end{tabular} & \begin{tabular}[c]{@{}r@{}}156.2\\ (154.90, 1.28)\end{tabular} & \begin{tabular}[c]{@{}r@{}}23.8\\ (20.83, 3.00)\end{tabular} & \begin{tabular}[c]{@{}r@{}}21.8\\ (20.31, 1.34)\end{tabular} & -                                                          & \multicolumn{1}{r|}{\begin{tabular}[c]{@{}r@{}}6.2\\ (4.93, 1.28)\end{tabular}} & \textbf{3.5-31.8$\times$} \\ \cline{1-1} \cline{3-9} 
\multicolumn{1}{l|}{Mistral-7b}                  & \multicolumn{1}{l|}{}                                                                                                                                                         & \begin{tabular}[c]{@{}r@{}}266.2\\ (264.93, 1.31)\end{tabular} & \begin{tabular}[c]{@{}r@{}}210\\ (208.41, 1.57)\end{tabular}   & -                                                            & \begin{tabular}[c]{@{}r@{}}21.5\\ (20.10, 0.41)\end{tabular} & -                                                          & \multicolumn{1}{r|}{\begin{tabular}[c]{@{}r@{}}6.4\\ (4.82, 1.57)\end{tabular}} & \textbf{3.4-41.6$\times$} \\ \hline
\multicolumn{2}{l|}{\textbf{Geo-mean (speedup)}}                                                                                                                                                 & \textbf{31.0$\times$}                                          & \textbf{19.6$\times$}                                          & \textbf{4.4$\times$}                                         & \textbf{3.4$\times$}                                         & \textbf{1.6$\times$}                                       & \multicolumn{1}{r|}{\textbf{-}}                                                 &                           \\ \hline
\multicolumn{1}{l|}{Qwen1.5-1.8B}                & \multicolumn{1}{l|}{\multirow{5}{*}{\begin{tabular}[c]{@{}l@{}}DroidTask: clock \\ prompt length: 656--827 tokens\\  output length: 1--5 tokens)\end{tabular}}}               & \begin{tabular}[c]{@{}r@{}}21\\ (20.84, 0.14)\end{tabular}     & \begin{tabular}[c]{@{}r@{}}10.4\\ (10.16, 0.25)\end{tabular}   & \begin{tabular}[c]{@{}r@{}}3.9\\ (3.25, 0.61)\end{tabular}   & -                                                            & -                                                          & \multicolumn{1}{r|}{\begin{tabular}[c]{@{}r@{}}1.4\\ (1.16, 0.25)\end{tabular}} & \textbf{2.80-15$\times$}  \\ \cline{1-1} \cline{3-9} 
\multicolumn{1}{l|}{Gemma-2B}                    & \multicolumn{1}{l|}{}                                                                                                                                                         & \begin{tabular}[c]{@{}r@{}}39.4\\ (39.11, 0.25)\end{tabular}   & \begin{tabular}[c]{@{}r@{}}16.5\\ (16.25, 0.21)\end{tabular}   & -                                                            & -                                                            & \begin{tabular}[c]{@{}r@{}}2.5\\ (2.30, 0.18)\end{tabular} & \multicolumn{1}{r|}{\begin{tabular}[c]{@{}r@{}}1.2\\ (0.97, 0.23)\end{tabular}} & \textbf{2.1-32.8$\times$} \\ \cline{1-1} \cline{3-9} 
\multicolumn{1}{l|}{Phi-2-2.7B}                  & \multicolumn{1}{l|}{}                                                                                                                                                         & \begin{tabular}[c]{@{}r@{}}46.6\\ (46.40, 0.24)\end{tabular}   & \begin{tabular}[c]{@{}r@{}}25\\ (24.62, 0.34)\end{tabular}     & \begin{tabular}[c]{@{}r@{}}7.4\\ (6.34, 1.06)\end{tabular}   & -                                                            & \begin{tabular}[c]{@{}r@{}}4.2\\ (3.98, 0.2)\end{tabular}  & \multicolumn{1}{r|}{\begin{tabular}[c]{@{}r@{}}3.1\\ (2.72, 0.34)\end{tabular}} & \textbf{1.4-15.0$\times$} \\ \cline{1-1} \cline{3-9} 
\multicolumn{1}{l|}{LlaMA-2-7B}                  & \multicolumn{1}{l|}{}                                                                                                                                                         & \begin{tabular}[c]{@{}r@{}}87.7\\ (87.38, 0.33)\end{tabular}   & \begin{tabular}[c]{@{}r@{}}60.4\\ (59.83, 0.63)\end{tabular}   & \begin{tabular}[c]{@{}r@{}}10.6\\ (8.58, 2.11)\end{tabular}  & \begin{tabular}[c]{@{}r@{}}11.1\\ (10.74, 0.37)\end{tabular} & -                                                          & \multicolumn{1}{r|}{\begin{tabular}[c]{@{}r@{}}4.8\\ (4.28, 0.53)\end{tabular}} & \textbf{2.2-18.3$\times$} \\ \cline{1-1} \cline{3-9} 
\multicolumn{1}{l|}{Mistral-7b}                  & \multicolumn{1}{l|}{}                                                                                                                                                         & \begin{tabular}[c]{@{}r@{}}122.3\\ (121.82, 0.51)\end{tabular} & \begin{tabular}[c]{@{}r@{}}68.6\\ (67.91, 0.65)\end{tabular}   & -                                                            & \begin{tabular}[c]{@{}r@{}}12\\ (11.64, 0.37)\end{tabular}   & -                                                          & \multicolumn{1}{r|}{\begin{tabular}[c]{@{}r@{}}4.9\\ (4.26, 0.65)\end{tabular}} & \textbf{2.4-25.0$\times$} \\ \hline
\multicolumn{2}{l|}{\textbf{Geo-mean (speedup)}}                                                                                                                                                                                 & \textbf{20.2$\times$}                                          & \textbf{10.8$\times$}                                          & \textbf{2.4$\times$}                                         & \textbf{2.4$\times$}                                         & \textbf{1.7$\times$}                                       & \multicolumn{1}{r|}{\textbf{-}}                                                 &                           \\ \hline
\multicolumn{1}{l|}{Qwen1.5-1.8B}                & \multicolumn{1}{l|}{\multirow{5}{*}{\begin{tabular}[c]{@{}l@{}}DroidTask: clock \\ prompt length: 505--645 tokens\\  output length: 3--5 tokens\end{tabular}}}                & \begin{tabular}[c]{@{}r@{}}16.2\\ (16.05, 0.14)\end{tabular}   & \begin{tabular}[c]{@{}r@{}}8.1\\ (7.83, 0.25)\end{tabular}     & \begin{tabular}[c]{@{}r@{}}3.1\\ (2.5, 0.61)\end{tabular}    & -                                                            & -                                                          & \multicolumn{1}{r|}{\begin{tabular}[c]{@{}r@{}}1.1\\ (0.90, 0.25)\end{tabular}} & \textbf{2.8-14.7$\times$} \\ \cline{1-1} \cline{3-9} 
\multicolumn{1}{l|}{Gemma-2B}                    & \multicolumn{1}{l|}{}                                                                                                                                                         & \begin{tabular}[c]{@{}r@{}}29.4\\ (29.14, 0.52)\end{tabular}   & \begin{tabular}[c]{@{}r@{}}12.3\\ (12.11, 0.20)\end{tabular}   & -                                                            & -                                                            & \begin{tabular}[c]{@{}r@{}}1.9\\ (1.71, 0.18)\end{tabular} & \multicolumn{1}{r|}{\begin{tabular}[c]{@{}r@{}}0.9\\ (0.72, 0.23)\end{tabular}} & \textbf{2.1-32.7$\times$} \\ \cline{1-1} \cline{3-9} 
\multicolumn{1}{l|}{Phi-2-2.7B}                  & \multicolumn{1}{l|}{}                                                                                                                                                         & \begin{tabular}[c]{@{}r@{}}35.4\\ (35.15, 0.23)\end{tabular}   & \begin{tabular}[c]{@{}r@{}}19\\ (18.65, 0.34)\end{tabular}     & \begin{tabular}[c]{@{}r@{}}5.9\\ (4.80, 1.05)\end{tabular}   & -                                                            & \begin{tabular}[c]{@{}r@{}}3.2\\ (3.01, 1.2)\end{tabular}  & \multicolumn{1}{r|}{\begin{tabular}[c]{@{}r@{}}2.4\\ (2.07, 0.34)\end{tabular}} & \textbf{1.3-14.8$\times$} \\ \cline{1-1} \cline{3-9} 
\multicolumn{1}{l|}{LlaMA-2-7B}                  & \multicolumn{1}{l|}{}                                                                                                                                                         & \begin{tabular}[c]{@{}r@{}}63.7\\ (63.38, 0.33)\end{tabular}   & \begin{tabular}[c]{@{}r@{}}43.9\\ (43.40, 0.52)\end{tabular}   & \begin{tabular}[c]{@{}r@{}}7.7\\ (5.06. 1.64)\end{tabular}   & \begin{tabular}[c]{@{}r@{}}8.2\\ (7.79, 0.37)\end{tabular}   & -                                                          & \multicolumn{1}{r|}{\begin{tabular}[c]{@{}r@{}}3.6\\ (3.10, 0.52)\end{tabular}} & \textbf{2.1-17.7$\times$} \\ \cline{1-1} \cline{3-9} 
\multicolumn{1}{l|}{Mistral-7b}                  & \multicolumn{1}{l|}{}                                                                                                                                                         & \begin{tabular}[c]{@{}r@{}}90.1\\ (89.60, 0.51)\end{tabular}   & \begin{tabular}[c]{@{}r@{}}50.6\\ (49.95, 0.65)\end{tabular}   & -                                                            & \begin{tabular}[c]{@{}r@{}}8.9\\ (8.56, 0.37)\end{tabular}   & -                                                          & \multicolumn{1}{r|}{\begin{tabular}[c]{@{}r@{}}3.8\\ (3.13, 0.65)\end{tabular}} & \textbf{2.3-23.7$\times$} \\ \hline
\multicolumn{2}{l|}{\textbf{Geo-mean (speedup)}}                                                                                                                                                                                 & \textbf{19.7$\times$}                                          & \textbf{10.5$\times$}                                          & \textbf{2.5$\times$}                                         & \textbf{2.3$\times$}                                         & \textbf{1.7$\times$}                                       & \multicolumn{1}{r|}{\textbf{-}}                                                 &                           \\ \hline

\multicolumn{1}{l|}{Qwen1.5-1.8B}                & \multicolumn{1}{l|}{\multirow{5}{*}{\begin{tabular}[c]{@{}l@{}} Persona-Chat \\ prompt length: 488--584 tokens\\  output length: 35--57 tokens\end{tabular}}}                & \begin{tabular}[c]{@{}r@{}}18.74\\ (16.54, 2.20)\end{tabular}   & \begin{tabular}[c]{@{}r@{}}11.86\\ (8.07, 3.79)\end{tabular}     & \begin{tabular}[c]{@{}r@{}}11.87\\ (2.58, 9.29)\end{tabular}    & -                                                            & -                                                          & \multicolumn{1}{r|}{\begin{tabular}[c]{@{}r@{}}6.72\\ (2.92, 3.80)\end{tabular}} & \textbf{1.8-2.8$\times$} \\ \cline{1-1} \cline{3-9} 
\multicolumn{1}{l|}{Gemma-2B}                    & \multicolumn{1}{l|}{}                                                                                                                                                         & \begin{tabular}[c]{@{}r@{}}30.9\\ (27.05, 3.82)\end{tabular}   & \begin{tabular}[c]{@{}r@{}}14.3\\(11.24, 3.02)\end{tabular}   & -                                                            & -                                                            & \begin{tabular}[c]{@{}r@{}}4.3\\ (1.59, 2.67)\end{tabular} & \multicolumn{1}{r|}{\begin{tabular}[c]{@{}r@{}}4.2\\ (0.67, 3.48)\end{tabular}} & \textbf{1.02-7.4$\times$} \\ \cline{1-1} \cline{3-9} 
\multicolumn{1}{l|}{Phi-2-2.7B}                  & \multicolumn{1}{l|}{}                                                                                                                                                         & \begin{tabular}[c]{@{}r@{}}36.1\\ (32.51, 3.57)\end{tabular}   & \begin{tabular}[c]{@{}r@{}}22.5\\ (15
.24, 3.22)\end{tabular}     & \begin{tabular}[c]{@{}r@{}}20.6\\ (4.43, 16.19)\end{tabular}   & -                                                            & \begin{tabular}[c]{@{}r@{}}5.9\\ (2.79, 3.06)\end{tabular}  & \multicolumn{1}{r|}{\begin{tabular}[c]{@{}r@{}}5.1\\ (1.91, 3.22)\end{tabular}} & \textbf{1.2-7.1$\times$} \\ \cline{1-1} \cline{3-9} 
\multicolumn{1}{l|}{LlaMA-2-7B}                  & \multicolumn{1}{l|}{}                                                                                                                                                         & \begin{tabular}[c]{@{}r@{}}64.2\\ (58.96, 5.20)\end{tabular}   & \begin{tabular}[c]{@{}r@{}}48.7\\ (40.37, 8.30)\end{tabular}   & \begin{tabular}[c]{@{}r@{}}45.5\\ (7.15, 38.23)\end{tabular}   & \begin{tabular}[c]{@{}r@{}}13.0\\ (7.25, 5.77)\end{tabular}   & -                                                          & \multicolumn{1}{r|}{\begin{tabular}[c]{@{}r@{}}11.2\\ (2.88, 8.30)\end{tabular}} & \textbf{1.2-5.7$\times$} \\ \cline{1-1} \cline{3-9} 
\multicolumn{1}{l|}{Mistral-7b}                  & \multicolumn{1}{l|}{}                                                                                                                                                         & \begin{tabular}[c]{@{}r@{}}90.5\\ (82.54, 8.0)\end{tabular}   & \begin{tabular}[c]{@{}r@{}}56.3\\ (46.02, 10.28)\end{tabular}   & -                                                            & \begin{tabular}[c]{@{}r@{}}13.7\\ (7.89, 5.76)\end{tabular}   & -                                                          & \multicolumn{1}{r|}{\begin{tabular}[c]{@{}r@{}}13.2\\ (2.88, 10.28)\end{tabular}} & \textbf{1.04-6.9$\times$} \\ \hline
\multicolumn{2}{l|}{\textbf{Geo-mean (speedup)}}                                                                                                                                                                                 & \textbf{10.5$\times$}                                          & \textbf{3.5$\times$}                                          & \textbf{3.1$\times$}                                         & \textbf{1.1$\times$}                                         & \textbf{1.1$\times$}                                       & \multicolumn{1}{r|}{\textbf{-}}                                                 &                           \\ \hline

\multicolumn{9}{l}{*\textit{LCPP} and \textit{PI} in the first row represent llama.cpp and PowerInfer-V2, respectively.}                                                                                                                                                                            
\end{tabular}
}
\label{tab:sec-eval-end-to-end}
\end{table*}

%% file: table-eval-accuracy.tex
\begin{table}[t]
    \caption{LLM capability accuracy on \sys and baselines. "SQ": SmoothQuant; "Int8()": LLM.Int8(); "Degrad.": accuracy degradation compared to FP16.}
    \label{tab:sec-eval-accuracy}
    \begin{subtable}[t]{0.48\textwidth}
        \resizebox{\textwidth}{!}{%
            \begin{tabular}{l|rrrrr|r}
                \toprule
                \textbf{LAMBADA}      & \textbf{FP16} & \textbf{SQ}      & \textbf{INT8()} & \textbf{K-Quant} & \textbf{Ours}   & \textbf{Ours Degrad.} \\
                \midrule
                Qwen1.5-1.8B          & 71.1\%        & 65.6\%           & 71.0\%          & 62.7\%           & 71.7\%          & \textbf{+0.6\%}       \\
                Gemma2-2B              & 59.6\%        & 45.8\%           & 59.2\%          & 56.9\%           & 59.4\%          & \textbf{-0.2\%}       \\
                Phi-2-2.7B             & 72.2\%        & 66.1\%           & 71.7\%          & 59.3\%           & 67.5\%          & \textbf{-4.7\%}       \\
                LlaMA-2-7B             & 87.5\%        & 71.9\%           & 88.0\%          & 15.6\%           & 86.3\%          & \textbf{-1.2\%}       \\
                Mistral-7b            & 84.8\%        & 51.2\%           & 85.3\%          & 23.9\%           & 84.1\%          & \textbf{-0.7\%}       \\
                \midrule
                \textbf{Avg. Degrad.} & -             & \textbf{-14.9\%} & \textbf{0\%}    & \textbf{-31.3\%} & \textbf{-1.2\%} &                       \\
                \bottomrule
            \end{tabular}


        }
    \end{subtable}
    \begin{subtable}[t]{0.48\textwidth}
        \resizebox{\textwidth}{!}{%
            \begin{tabular}{l|rrrrr|r}
                \toprule
                \textbf{HellaSwag}    & \textbf{FP16} & \textbf{SQ}     & \textbf{INT8()} & \textbf{K-Quant} & \textbf{Ours}   & \textbf{Ours Degrad.} \\
                \midrule
                Qwen1.5-1.8B          & 43.8\%        & 40.9\%          & 43.5\%          & 44.3\%           & 43.8\%          & \textbf{0\%}          \\
                Gemma2-2B              & 46.5\%        & 43.8\%          & 46.1\%          & 45.4\%           & 47.3\%          & \textbf{+0.8\%}       \\
                Phi-2-2.7B             & 48.2\%        & 46.2\%          & 47.7\%          & 47.6\%           & 46.9\%          & \textbf{-1.3\%}       \\
                LlaMA-2-7B             & 52.8\%        & 44.4\%          & 53.1\%          & 50.5\%           & 53.5\%          & \textbf{+0.7\%}       \\
                Mistral-7b            & 57.4\%        & 44.9\%          & 57.9\%          & 57.0\%           & 57.0\%          & \textbf{-0.4\%}       \\
                \midrule
                \textbf{Avg. Degrad.} & -             & \textbf{-5.7\%} & \textbf{-0.1\%} & \textbf{-0.8\%}  & \textbf{-0.0\%} &                       \\
                \bottomrule
            \end{tabular}


        }
    \end{subtable}

    \begin{subtable}[t]{0.48\textwidth}
        \resizebox{\textwidth}{!}{%
            \begin{tabular}{l|rrrrr|r}
                \toprule
                \textbf{WinoGrande}   & \textbf{FP16} & \textbf{SQ}     & \textbf{INT8()} & \textbf{K-Quant} & \textbf{ours}   & \textbf{Ours Degrad.} \\
                \midrule
                Qwen1.5-1.8B          & 58.3\%        & 51.0\%          & 58.2\%              & 59.0\%           & 59.3\%          & \textbf{+1.0\%}          \\
                Gemma2-2B              & 58.3\%        & 54.8\%          & 59.0\%              & 58.5\%           & 59.5\%          & \textbf{+1.2\%}          \\
                Phi-2-2.7B             & 72.2\%        & 68.9\%          & 72.4\%              & 72.5\%           & 70.2\%          & \textbf{-2.0\%}          \\
                LlaMA-2-7B             & 65.2\%        & 56.9\%          & 66.2\%              & 67.4\%           & 65.1\%          & \textbf{-0.1\%}          \\
                Mistral-7b            & 73.5\%        & 59.1\%          & 73.3\%              & 73.5\%           & 73.1\%          & \textbf{-0.4\%}          \\
                \midrule
                \textbf{Avg. Degrad.} & -             & \textbf{-7.4\%} & \textbf{+0.3\%}     & \textbf{+0.7\%}  & \textbf{-0.1\%} &                          \\
                \bottomrule
            \end{tabular}


        }
    \end{subtable}
    \begin{subtable}[t]{0.48\textwidth}
        \resizebox{\textwidth}{!}{%
            \begin{tabular}{l|rrrrr|r}
                \toprule
                \textbf{OpenBookQA}   & \textbf{FP16} & \textbf{SQ}     & \textbf{INT8()} & \textbf{K-Quant} & \textbf{ours}   & \textbf{Ours Degrad.} \\
                \midrule
                Qwen1.5-1.8B          & 28.8\%        & 23.0\%          & 28.5\%              & 28.0\%           & 26.6\%          & \textbf{-2.2\%}          \\
                Gemma2-2B              & 33.7\%        & 28.0\%          & 34.2\%              & 33.0\%           & 38.4\%          & \textbf{+4.7\%}          \\
                Phi-2-2.7B             & 41.0\%        & 35.9\%          & 40.2\%              & 39.5\%           & 37.7\%          & \textbf{-3.3\%}          \\
                LlaMA-2-7B             & 32.7\%        & 25.0\%          & 32.0\%              & 31.5\%           & 31.1\%          & \textbf{-1.6\%}          \\
                Mistral-7b            & 39.4\%        & 25.6\%          & 39.3\%              & 37.9\%           & 39.3\%          & \textbf{-0.1\%}          \\
                \midrule
                \textbf{Avg. Degrad.} & -             & \textbf{-7.6\%} & \textbf{-0.3\%}     & \textbf{-1.1\%}  & \textbf{-0.5\%} &                          \\
                \bottomrule
            \end{tabular}


        }
    \end{subtable}

    \begin{subtable}[t]{0.48\textwidth}
        \resizebox{\textwidth}{!}{%
            \begin{tabular}{l|rrrrr|r}
                \toprule
                \textbf{MMLU}         & \textbf{FP16} & \textbf{SQ}     & \textbf{INT8()} & \textbf{K-Quant} & \textbf{ours}   & \textbf{Ours Degrad.} \\
                \midrule
                Qwen1.5-1.8B          & 29.7\%        & 27.9\%          & 29.1\%              & 29.8\%           & 30.8\%          & \textbf{+1.1\%}          \\
                Gemma2-2B              & 35.7\%        & 32.1\%          & 35.1\%              & 35.1\%           & 36.4\%          & \textbf{+0.7\%}          \\
                Phi-2-2.7B             & 35.4\%        & 35.3\%          & 35.6\%              & 35.7\%           & 36.7\%          & \textbf{+1.3\%}          \\
                LlaMA-2-7B             & 37.8\%        & 29.2\%          & 38.1\%              & 34.4\%           & 36.9\%          & \textbf{-0.9\%}          \\
                Mistral-7b            & 42.1\%        & 30.9\%          & 41.4\%              & 42.3\%           & 41.0\%          & \textbf{-1.1\%}          \\
                \midrule
                \textbf{Avg. Degrad.} & -             & \textbf{-5.1\%} & \textbf{-0.3\%}     & \textbf{-0.7\%}  & \textbf{+0.2\%} &                          \\
                \bottomrule
            \end{tabular}
        }
    \end{subtable}

\end{table}







%% file: figure-trade-off-speed-accuracy.tex
\begin{figure}
    \centering
    \begin{subfigure}[t]{0.23\textwidth}
        \includegraphics[width=\textwidth]{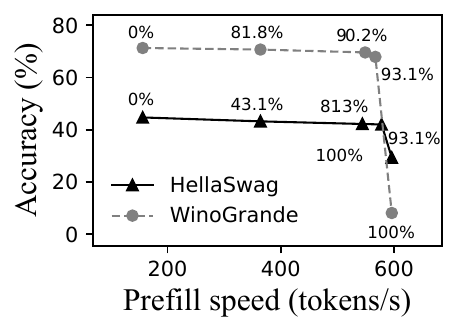}
        \caption{Qwen1.5-1.8B}
    \end{subfigure}
    \begin{subfigure}[t]{0.23\textwidth}
        \includegraphics[width=\textwidth]{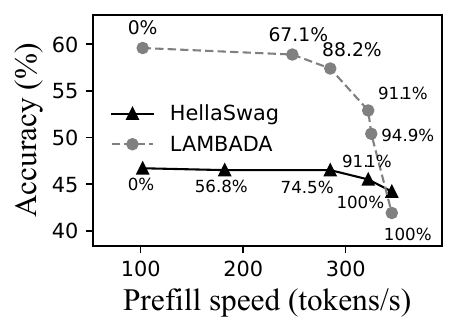}
        \caption{Gemma-2B}
    \end{subfigure}
        \caption{\revision{Generation speed and accuracy across various outlier pruning rates. The text positioned above or below the plots represents the corresponding pruning rates.}}
    \label{fig:sec-eval-trade-off-accuracy-speed}
\end{figure}

%% file: figure-eval-memory.tex
\begin{figure}[t]
    \centering
    \includegraphics[width=0.46\textwidth]{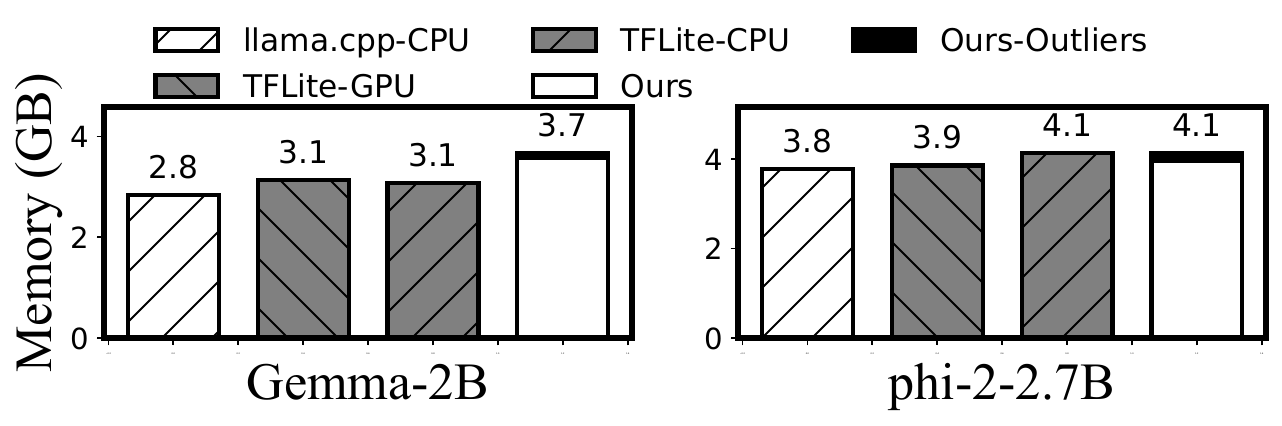}
    \caption{Memory consumption of different baselines (prompt length=512).}
    \label{fig:sec-eval-memory}
\end{figure}

%% file: figure-eval-gpu.tex
\begin{figure}
    \centering
    \begin{subfigure}[t]{0.23\textwidth}
        \includegraphics[width=\textwidth]{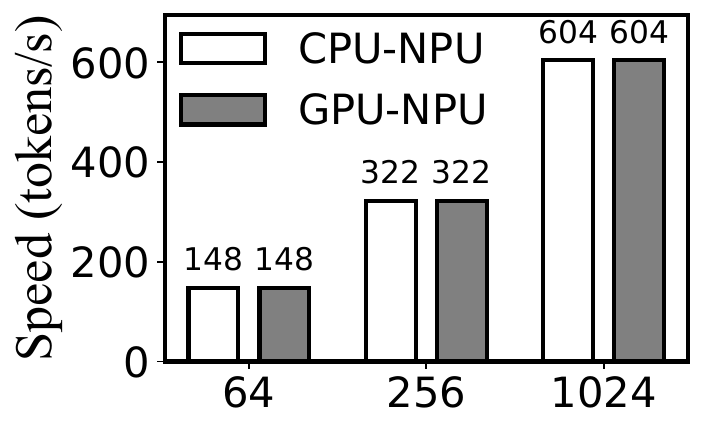}
        \caption{Prefill speed under different prompt length}
    \end{subfigure}
    \begin{subfigure}[t]{0.23\textwidth}
        \includegraphics[width=\textwidth]{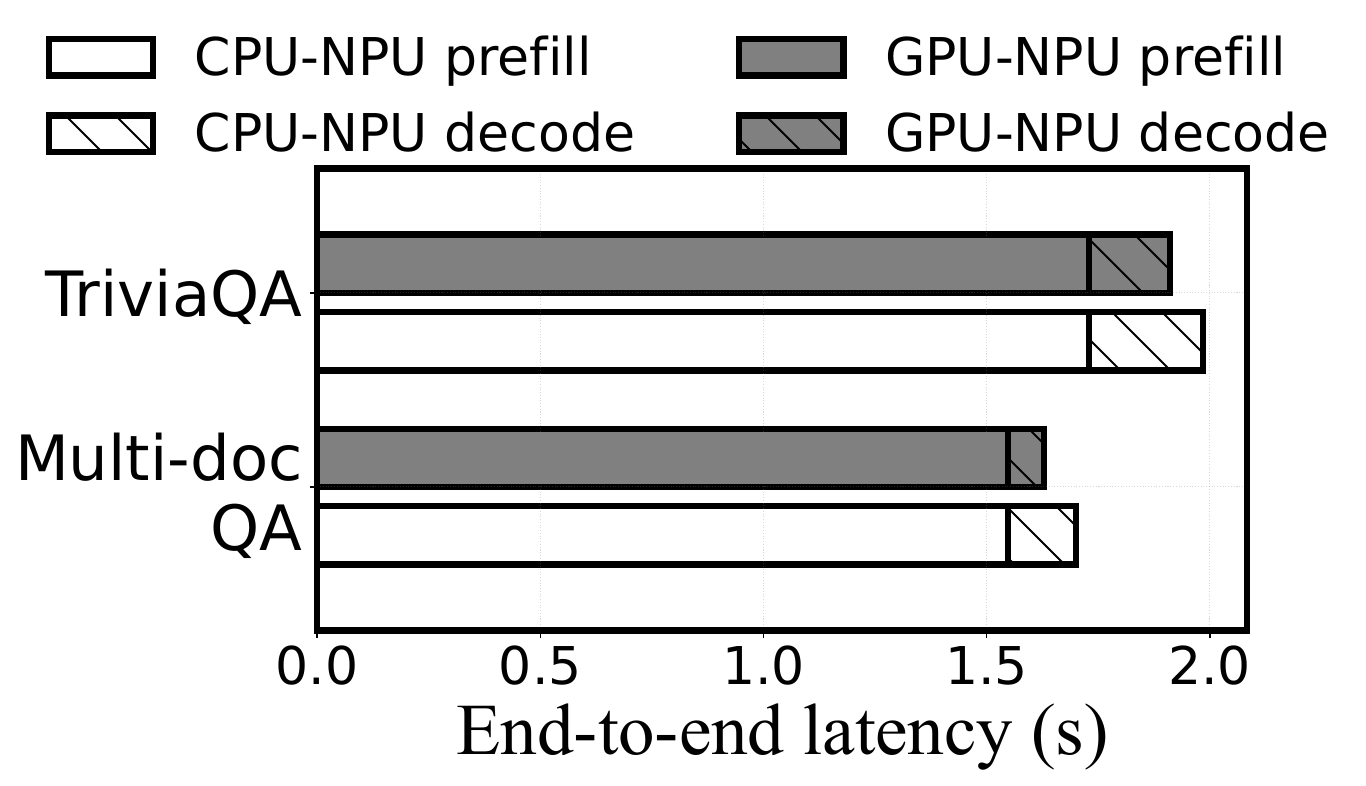}
        \caption{End-to-end latency on the Longbench dataset}
    \end{subfigure}
        \caption{\revision{Prefill speed and end-to-end latency comparison between CPU-NPU coordination and GPU-NPU coordination.}}
    \label{fig:sec-eval-gpu}
\end{figure}

%% file: figure-eval-ablation.tex
\begin{figure}
    \centering
    \includegraphics[width=0.48\textwidth]{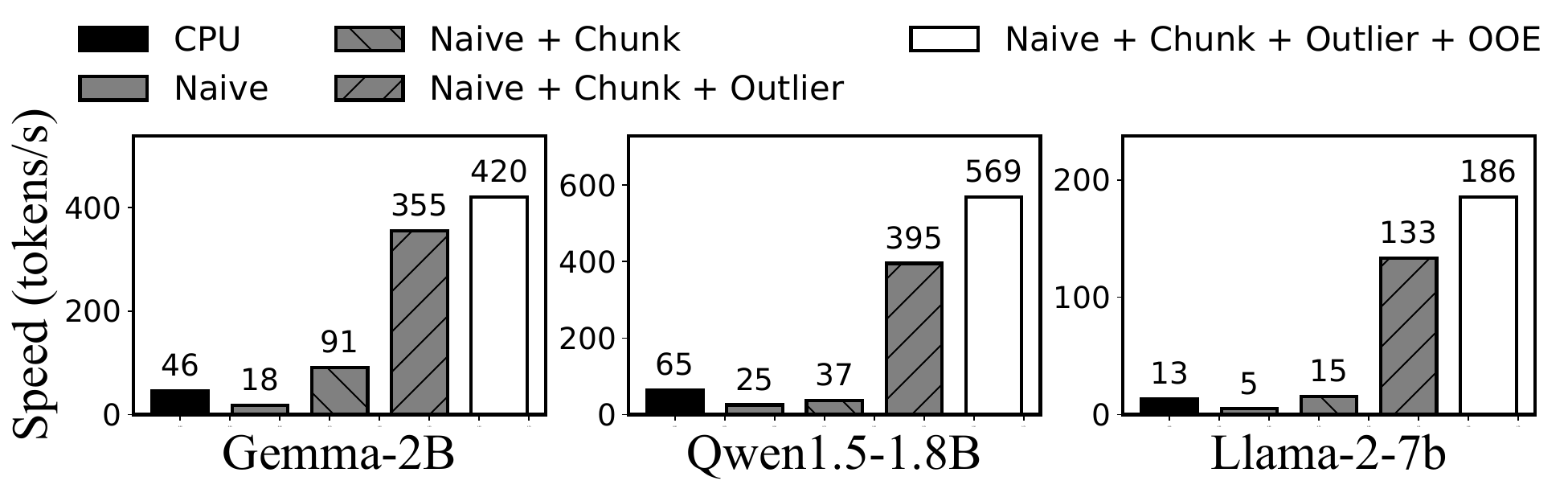}
    \caption{Ablation study of \sys (prompt length=512).}
    \label{fig:sec-eval-ablation}
\end{figure}

%% file: sec-discuss.tex
\section{Discussion and Future Work} \label{sec-discussion}
\textbf{Advantages of NPUs over GPUs.}
\revision{
Typically, both mobile GPUs and NPUs serve as hardware accelerators for on-device DNN inference~\cite{lane2016deepx,xu2022mandheling}.
However, as mobile GPUs are already extensively used for graphic rendering on devices, relying on them for LLM inference can cause significant resource contention. 
In contrast, mobile NPUs, which are dedicated exclusively to deep learning tasks, have relatively lighter workloads, leaving ample time to run LLM inference.
Besides, as elaborated in Section~\ref{sec-bg-npu}, mobile NPUs provide greater computational capacity, enhanced energy efficiency, and increased availability compared to mobile GPUs, making them more suitable for on-device LLM inference.}

\noindent \textbf{\sys's technique stability.}
\revision{
Despite the recent trend towards supporting FP16 in advanced NPUs~\cite{Meteor-Lake,AMD-Strix-Point,8gen3}, \sys remains essential for NPU-based prefilling acceleration for two reasons: (1) Integer operations are expected to maintain dominance in mobile NPU-side LLM inference due to their superior computation performance, memory efficiency, and energy savings, while CPU/GPUs serve as supplementary units for enhancing accuracy. Therefore, \sys's three INT8-related techniques, which include co-scheduling across CPU/GPU and NPU to achieve high accuracy and efficiency, will continue to have a significant impact in this domain.
(2) \sys is capable of empowering legacy devices with on-device LLM capabilities, democratizing access to LLMs.
}

\noindent \textbf{Future hardware design implications.}
\revision{
\sys's performance could be further enhanced with the following hardware optimizations:
(1) \textit{Dynamic shape-aware optimization}: Hardware capable of dynamically reconfiguring to accommodate varying input and output shapes would markedly improve the acceleration of LLM inference.
(2) \textit{Increased data cache size}: Mobile NPUs generally enhance inference performance by loading an entire layer's weights into the data cache. Increasing the NPU cache size to align with LLM weight sizes is crucial for optimal performance.
(3) \textit{Mixed-precision operands in computing units}: Mobile NPUs typically support only uniform precision operands, necessitating additional data format conversions to perform FP16 * INT8 operations. This overhead can be eliminated by enabling support for mixed-precision operands in computing units.}

\noindent \textbf{\sys with mobile GPUs.}
\revision{
Our current implementation relies solely on mobile CPUs for floating-point operators and the decoding phase due to ease of implementation. However, the design of \sys is compatible with both mobile CPUs and GPUs. 
We plan to incorporate GPU-NPU coordination in future developments. The primary engineering efforts in integrating mobile GPUs into \sys involve configuring the GPU backend and implementing the necessary operators.
With mobile GPUs, \sys can significantly reduce end-to-end inference latency, as elaborated in Section~\ref{sec:eval-gpu}. }


%% file: sec-related.tex
\section{Related work}
\textbf{On-device LLM optimization.}
To reduce the substantial memory consumption of on-device LLM inference, various compression techniques have been proposed, including quantization and knowledge distillation~\cite{yao2022zeroquant, frantar2022gptq, guan2021cocopie, niu2020patdnn, wang2020minilm, huynh2017deepmon, you2022speechmoe2,pope2023efficiently,mllm-npu,xu2024survey,yi2023edgemoe,cai2023efficient,xu2018deepcache,LiuYiSCIS23}. To minimize on-device LLM computation, researchers have introduced \textit{token pruning}~\cite{wang2021spatten, cai2022enable, kim2022learned, rao2021dynamicvit, bolya2022token}, which prunes unnecessary tokens during the inference process.
Speculative decoding, a method that accelerates token generation by offloading tasks to a smaller LLM, has been widely adopted in open-source frameworks~\cite{llama-cpp, FastTransformer} and extensively researched~\cite{kim2023big, miao2023specinfer, yang2023predictive, he2023rest, fu2024break, cai2024medusa,xu2023llmcad}. As a system optimization, \sys is orthogonal and compatible with these algorithm-level optimizations.

\noindent \textbf{On-chip offloading for ML.}
This has been thoroughly studied to enable faster DNN inference by leveraging heterogeneous mobile processors like GPUs and NPUs~\cite{kim2019mulayer, zeng2021energy, ha2021accelerating, lane2016deepx, han2019mosaic, zhang2020mobipose, lee2019mobisr, georgiev2014dsp, xu2022mandheling, xue2024powerinfer,niu2024smartmem,niu2024sod,hsu2023simultaneous,xu2024fwdllm,xu2023niagara,LiuSICS24}. MobiSR~\cite{lee2019mobisr} utilizes mobile NPUs to speed up super-resolution computation.
However, these methods do not address LLM-specific features and are unsuitable for on-device LLM scenarios. 
The most relevant work is PowerInfer-V2~\cite{xue2024powerinfer}, which also utilizes mobile NPUs for the prefilling.
But, it does not comprehensively analyze the challenges of on-device NPU offloading for LLM prefilling or incorporate techniques to fully harness NPU capability.
Our experiments demonstrate that our approach is up to 5.32$\times$ more efficient in the prefill stage. \sys is inspired by these efforts and is the first LLM inference framework with efficient on-device NPU offloading.

\noindent \textbf{Mobile NPU execution optimization.}
With the growing importance of mobile NPUs in smartphones, significant efforts have been made to optimize their execution efficiency~\cite{root2023fast, thomas2024automatic, niu2022gcd, vocke2017extending, ragan2013halide, franchetti2008generating}. Pitchfork~\cite{root2023fast} defines a portable fixed-point intermediate representation to optimize fixed-point execution efficiency. Isaria~\cite{thomas2024automatic} proposes a framework for automatically generating vectorizing compilers for DSP architectures, creating efficient operator codes for mobile NPUs.
As a system framework, \sys is orthogonal and can leverage them to generate more efficient operator libraries as execution backend, further boosting performance.




%% file: sec-conclusion.tex
\section{Conclusions}\label{sec:conclusions}
This paper has proposed \sys, the first LLM inference system utilizing on-device NPU offloading to reduce prefill latency and energy consumption.
\sys has incorporated novel techniques: chunk-sharing graph, shadow outlier execution and out-of-order subgraph execution to enhance NPU offloading efficiency.
Extensive experiments have demonstrated \sys to show its superior performance benefits, e.g, up to 43.6$\times$ speedup and 59.5$\times$ energy savings.

%% file: sec-ack.tex
\section{Acknowledgement}
This work was supported by the National Natural Science Foundation of China under grant number 62325201 and sponsored by the Huawei University Joint Research Program. 

%% file: sec-ae.tex
\clearpage
\appendix
\section{Artifact Appendix}

\subsection{Abstract}

The artificial evaluations primarily present the prefill performance results detailed in Section 4.2 and the inference accuracy evaluation results discussed in Section 4.4. These results can be found in the \textit{performance\_results} and \textit{accuracy\_results} directories in our GitHub repository\footnote{\url{https://doi.org/10.5281/zenodo.14392760}}. 
To conduct these evaluations, please refer to the README.md file in the repository. 
Each evaluation directory in the repository contains detailed instructions for installing and executing evaluations for \sys and all baseline methods. This includes guidance on environment setup, LLM and dataset downloads, source code compilation and execution, and clarity on results. 
We recommend performing the accuracy evaluation first, followed by the prefill performance test. For optimal results, use a Redmi K70 Pro 24G for on-device LLM inference and an NVIDIA A100 40G GPU server for testing, as these setups have been validated by the authors. Most experiments utilize Qwen1.5-1.8B as a representative example to demonstrate the functionality and evaluation results of \sys. Users have the flexibility to tailor their own LLM models by adjusting the code, if needed. Once this artifact evaluation is accepted, we will archive the evaluation results.

\subsection{Artifact check-list (meta-information)}

{\small
\begin{itemize}
  \item {\bf Compilation: Support compilation from users.}
  \item {\bf Binary: Compiled binaries are available.}
  \item {\bf Model: Qwen1.5-1.8B-Chat is used as demonstration model. Gemma-2B, Phi-2-2.7B, LLaMA-2-7B, and Mistral-7B are also supported.}
  \item {\bf Dataset: HellaSwag~\cite{zellers2019hellaswag} and LAMBADA~\cite{lambada} are used as demonstration datasets. OpenBookQA~\cite{OpenBookQA2018}, WinoGrande~\cite{ai2:winogrande}, and MMLU~\cite{hendryckstest2021} are also supported.}
  \item {\bf Server Hardware: NVIDIA-A100 40G}
  \item {\bf Device Hardware: Redmi K70 Pro 24G}
  \item {\bf Execution: C++, Python and bash scripts}
  \item {\bf Metrics: inference accuracy and prefill speed}
  \item {\bf Output: log in terminal}
  \item {\bf Experiments: Section 4.2 and Section 4.4}
  \item {\bf How much disk space required (approximately)?: 200GB}
  \item {\bf How much time is needed to prepare workflow (approximately)?: 5 hours}
  \item {\bf How much time is needed to complete experiments (approximately)?: 2 hours}
  \item {\bf Publicly available?: Yes}
  \item {\bf Code licenses (if publicly available)?: MIT}
  \item {\bf Data licenses (if publicly available)?: MIT}
  \item {\bf Workflow automation framework used?: mllm~\cite{mllm}}
  \item {\bf Archived (provide DOI)?: \url{https://doi.org/10.5281/zenodo.14392760}.}
\end{itemize}
}

\subsection{Description}

\subsubsection{How to access}

Our artificial evaluation is publicly accessible at 

\noindent \url{https://doi.org/10.5281/zenodo.14392760}.

\subsubsection{Hardware dependencies}
For the accuracy results discussed in Section 4.4, reviewers will require access to a GPU server equipped with an NVIDIA A100 40GB GPU, x86 CPU cores, and preferably over 200GB of CPU memory. The server should operate on Ubuntu 20.04 OS.

To assess the performance results in Section 4.2, reviewers will need a Redmi K70 Pro with 24GB of memory.

\subsubsection{Software dependencies}
For the accuracy results outlined in Section 4.4, the server should have CUDA 12.5 installed and a Conda virtual environment with Python version 3.8 or higher.

\subsubsection{Datasets}
The datasets used for accuracy results include HellaSwag~\cite{zellers2019hellaswag}, LAMBADA, OpenBookQA~\cite{OpenBookQA2018}, WinoGrande~\cite{ai2:winogrande}, and MMLU~\cite{hendryckstest2021}. All datasets can be automatically downloaded using our code from Hugging Face~\cite{Hugging-Face}.

\subsubsection{Models}
We demonstrate \sys's functionality on devices using Qwen1.5-1.8B. Users may also perform additional evaluations by integrating customized models like Gemma-2B, Phi-2-2.7B, LLaMA-2-7B, and Mistral-7B.

\subsection{Installation}
To obtain the accuracy results discussed in Section 4.4, users should begin by using Conda to install the Python virtual environment specified in \textit{accuracy\_results/environment.yml}. This environment includes all the necessary dependencies to reproduce the accuracy results of \sys.
When evaluating the baseline methods, particularly SmoothQuant~\cite{xiao2023smoothquant}, users need to install the SmoothQuant environment as detailed in \textit{accuracy\_results/baselines/README.md}.

For the performance results outlined in Section 4.2, users are advised to refer to and follow the installation instructions provided in each respective \textit{README.md} file within the \textit{performance\_results} directory.

\subsection{Experiment workflow}
Initially, users should perform the accuracy results evaluation (\textit{accuracy\_results}) to generate the quantized model necessary for on-device inference with \sys (\textit{accuracy\_results/ llm.npu/examples/act\_scales\_try/llmnpu\_get\_int8\_weights \_finalmodel \_int8bias\_ns.py}). Subsequently, using the generated model, users should proceed with the performance results evaluation (\textit{performance\_results}).

\subsection{Evaluation and expected results}
We offer two primary types of evaluation results: accuracy evaluation results for Section 4.4 and prefill performance evaluation results for Section 4.2. These results contribute to the data reported in Table 6 and illustrated in Figure 14.

\subsection{Methodology}
Availability, Functional, and Reproducibility